\documentclass[10pt,twocolumn,letterpaper]{article}

\usepackage{wacv}
\usepackage{times}
\usepackage{epsfig}
\usepackage{graphicx}
\usepackage{amsmath}
\usepackage{amssymb}
\usepackage{booktabs}
\usepackage{arydshln}
\usepackage{xcolor}
\usepackage{multirow}
\def\wacvPaperID{854} 

\wacvapplicationstrack 

\wacvfinalcopy 

\ifwacvfinal
\usepackage[breaklinks=true,bookmarks=false]{hyperref}
\else
\usepackage[pagebackref=true,breaklinks=true,colorlinks,bookmarks=false]{hyperref}
\fi

\begin{document}

\title{Exploring Semantic Consistency in Unpaired Image Translation to Generate Data for Surgical Applications}


\author{
Danush Kumar Venkatesh\textsuperscript{1,2,3}, Dominik Rivoir\textsuperscript{1,4}, Micha Pfeiffer\textsuperscript{1}, Fiona Kolbinger\textsuperscript{1,3},\\
Marius Distler\textsuperscript{3}, Jürgen Weitz\textsuperscript{3,4}, Stefanie Speidel\textsuperscript{1,2,3,4}
\and
\small{\textsuperscript{1}NCT/UCC Dresden},
\textsuperscript{2}SECAI, TU Dresden,
\textsuperscript{3}Faculty of medicine \& University Hospital TU Dresden,
\textsuperscript{4}CeTI, TU Dresden\\
{\tt\small \{danushkumar.venkatesh, dominik.rivoir, micha.pfeiffer, stefanie.speidel\}@nct-dresden.de}\\
{\tt\small \{fiona.kolbinger, marius.distler, juergen.weitz\}@uniklinikum-dresden.de}
}

\maketitle
\thispagestyle{empty}

\begin{abstract}
   
In surgical computer vision applications, obtaining labeled training data is challenging due to data-privacy concerns and the need for expert annotation. Unpaired image-to-image translation techniques have been explored to automatically generate large annotated datasets by translating synthetic images to the realistic domain. However, preserving the structure and semantic consistency between the input and translated images presents significant challenges, mainly when there is a distributional mismatch in the semantic characteristics of the domains. This study empirically investigates unpaired image translation methods for generating suitable data in surgical applications, explicitly focusing on semantic consistency. We extensively evaluate various state-of-the-art image translation models on two challenging surgical datasets and downstream semantic segmentation tasks. We find that a simple combination of structural-similarity loss and contrastive learning yields the most promising results. Quantitatively, we show that the data generated with this approach yields higher semantic consistency and can be used more effectively as training data. The code is availbale at \url{https://gitlab.com/nct_tso_public/constructs}.
\end{abstract}

\section{Introduction}
The field of surgical data science has witnessed a resurgence in recent years, propelled by rapid advancements in data science methodologies and deep learning (DL) techniques~\cite{maier2022surgical}. Meanwhile, clinical innovations such as robot- and computer-assisted surgical systems have revolutionized minimally-invasive surgery over the last decade by providing intraoperative guidance and decision support~\cite{haidegger2022robot, maier2017surgical, bodenstedt2020artificial}.

However, the field encounters a significant constraint by the limited access to large annotated datasets, which impedes the potential for training large and powerful models~\cite{maier2017surgical, maier2022surgical}. Multiple challenges contribute to this limitation, including the technical complexities in acquiring patient data directly from the operating room~\cite{hager2020surgical}, legal regulations on data sharing, and the substantial costs involved in expert labeling, given the restricted availability of domain specialists (i.e., surgical professionals). 
One potential solution to overcome these challenges is adopting synthetic training data generated through computer simulations~\cite{sharan2021mutually, pfeiffer2019generating,yoon2022surgical, rivoir2021long}. Synthetic data presents the advantage of automatically generating substantial volumes of fully labeled data. Nonetheless, enforcing real-world characteristics in such synthetic datasets can be a significant hurdle.

\begin{figure}
\begin{center}
\includegraphics[width=0.4\textwidth]{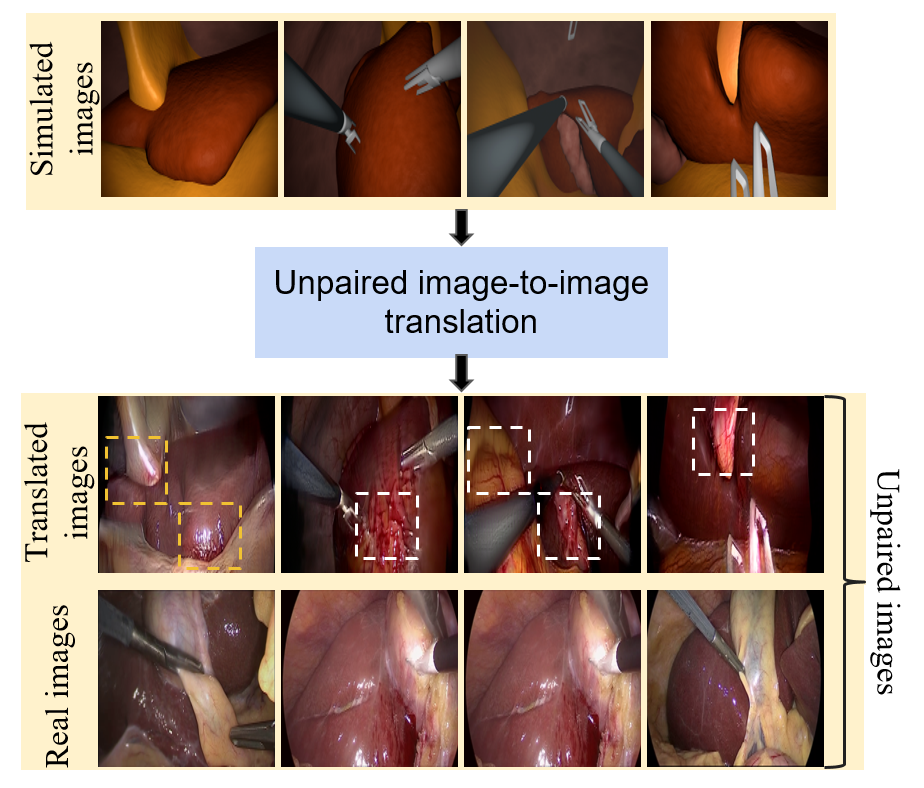}
\end{center}
   \caption{Generation of realistic data from synthetic surgical images with unpaired image translation method. The semantic mismatch between domains can lead to inconsistent translations, like blood texture (red color) getting mapped onto different structures (highlighted in white boxes). Some regions with consistent semantic translation are indicated in yellow boxes.}
\label{fig:intro}
\end{figure}

Image-to-image translation (I2I) methods are generative modeling techniques that have gained popularity for translating images between different domains. Within the field of data generation, the applicability of paired image translation methods~\cite{isola2017image, liu2019few} is limited. Conversely, unpaired image translation methods~\cite{zhu2017unpaired}, which do not require corresponding image pairs, have emerged as promising solutions for various computer vision tasks. These methods are employed in tasks such as translating synthetic images into realistic ones, performing style transfer, and adapting images across different domains~\cite{pfeiffer2019generating, benaim2017one, huang2018multimodal, liu2017unsupervised, zhang2017stackgan, choi2018stargan, hoffman2018cycada, dosovitskiy2016generating, ledig2017photo}. 
Overall unpaired image translation methods are suitable for surgical applications, but they face challenges in preserving contextual and semantic details across the domains. 

In practice, translation methods aim to align the image statistics between the two domains. In addition to the difference in image distributions, semantic variations in distributions also exist, which is commonly referred to as "\emph{unmatched semantic statistics}"~\cite{jia2021semantically} and poses a critical problem in preserving the semantics during translation. As displayed in Figure~\ref{fig:intro}, when faced with unmatched semantic distributions, attempting to align the distributions between translated and target images forcibly can result in spurious solutions, where semantic information is distorted~\cite{guo2022alleviating, jia2021semantically}.

Several approaches have been proposed to preserve semantics during image translations and mitigate semantic distortion. However, these methods might require additional supervision or pre-trained models~\cite{taigman2016unsupervised, hoffman2018cycada}. Alternatively, some approaches are excessively restrictive, tailored to specific datasets, and prone to introducing artifacts~\cite{benaim2017one, zhang2019harmonic}. Furthermore, methods based on mutual information between the images~\cite{guo2022alleviating} and optimization of robustness loss~\cite{jia2021semantically} have also been explored to mitigate this issue. 

In real surgical scenarios, an additional challenge arises from the variations in lighting conditions, which may not be adequately reflected in existing baseline datasets~\cite{cordts2016cityscapes, isola2017image, chu2017cyclegan}. While synthetic images can incorporate such parameters, creating such a realistic environment takes time and effort. The central idea would be to develop simple virtual scenarios and utilize DL approaches to enhance realism. Semantic consistency can be affected when such variations exist and addressing these short comings is essential as without doing so, the generated data lacks practical utility for subsequent training of models (Section ~\ref{sec:results}).

 To the best of our knowledge, this study represents the first comprehensive investigation of unpaired image translation techniques to generate data in the context of surgical applications. We summarize our contribution as follows.
\begin{itemize} 
\item We empirically analyze various methods for unpaired image translation by assessing both the semantic consistency of the translated images and their utility as training data in diverse downstream tasks. 
\item We tackle the underexplored problem of creating \emph{semantic consistent datasets with annotations}. We focus on translating synthetic anatomical images into realistic surgical images on datasets from minimally-invasive surgeries, namely, cholecystectomy and gastrectomy. 
    \item Guided by our analysis, we define a novel combination of an image quality assessment metric~\cite{wang2003multiscale} as a loss function with the contrastive learning framework~\cite{park2020contrastive} as a simple yet effective modification to tackle the challenge of semantic distortion.
    \item We found promising results that this method is more effective than many of the existing unpaired translation methods, highlighting its effectiveness in maintaining semantic consistency.
\end{itemize}


\section{Related Work \& Background}
\textbf{Image-to-Image translation}.
The objective is to generate images in a desired target domain while preserving the structure and semantics of the input. Generative adversarial networks (GANs)~\cite{goodfellow2020generative}  have proven to be a powerful approach for image translation, learning the mapping between input and output images. 
While alternative methods like diffusion models and variational autoencoders (VAEs) have been proposed, research is still in a nascent stage, and they are mainly employed in supervised conditional settings or purely generative contexts~\cite{saharia2022palette, wang2022pretraining, wang2022semantic, rombach2022high, su2022dual}. GANs still remain a prime choice for unpaired image translation tasks. 
In the case of unpaired translation, a technique called cycle consistency~\cite{zhu2017unpaired} was introduced, which seeks to learn the reverse mapping between different domains by leveraging reconstruction loss. Various approaches have been proposed to address multi-modal and domain translations, focusing on disentangling images' content and style information in distinct spaces~\cite{liu2017unsupervised, huang2018multimodal, pfeiffer2019generating, choi2018stargan, lee2018diverse, zhu2017toward}. Although these approaches effectively exploit cycle consistency, they often rely on the assumption of a bijective relationship between domains, which can be overly restrictive. Achieving perfect reconstruction becomes challenging, mainly when a semantic mismatch exists between the domains~\cite{park2020contrastive, jia2021semantically}. 

To address this limitation, one-sided translation methods have been proposed as alternatives to cycle consistency. For instance, GcGAN~\cite{fu2019geometry} incorporates an equivariance constraint for geometric image transformations, while DistanceGAN~\cite{tran2018dist} enforces consistency regularization based on distances between the images. Similarly, HarmonicGAN~\cite{zhang2019harmonic} enforces visual similarities between domains, and TraVeLGAN~\cite{amodio2019travelgan} preserves the arithmetic properties of embedding vectors. Attention-based techniques~\cite{tang2021attentiongan, wang2021instance, alami2018unsupervised, shrivastava2017learning, wang2018high} have also been proposed to maintain semantic coherence before and after the translation process. Efforts such as~\cite{dosovitskiy2016generating, johnson2016perceptual, mechrez2018contextual} have been made to minimize the perceptual or content loss by utilizing a pre-trained VGG model to decrease the content disparity between the domains. However, this approach is computationally expensive and lacks adaptability to the available data. A contrastive learning-based image translation method was proposed in the CUT~\cite{park2020contrastive} model. 

\textbf{Semantic robustness via losses}.
Recently, two approaches were proposed to minimize semantic distortion during translation. SRUNIT~\cite{jia2021semantically}, based on CUT, proposed a semantic robustness loss that is optimized between the input features of the domain $\mathcal{X}$ with the perturbated variant of the same. The intuition is that the semantics of the output should remain invariant to any perceptual (image level) changes in the input. The loss is defined as,
\begin{equation}
\begin{aligned}
\mathcal{L}_{r}=\mathbb{E}_x\left[\frac{1}{\left\|\tau_r\right\|_2} \|\right. & F_r\left(G_{f}^r(x)\right)- \\
& \left.F_r\left(G_{\mathcal{XY}}^r\left(\left(G_f^r(x)+\tau_r\right)\right)\right) \|_2\right],
\end{aligned}
\end{equation}
where  $F_r$ indicates the feature extractor, $G_f^r$ is the feature extracted from the $f^th$ layer of the generative model $G$ and $\tau$ is the perturbation parameter. Similarly, a structural consistency constraint (SCC)~\cite{guo2022alleviating} was proposed to maintain the semantics. The color randomness in the pixel values of the images before and after the translation was reduced by exploiting mutual information. The SCC loss is defined as, 
\begin{equation}
\mathcal{L}_{S C C}=\frac{1}{N} \sum_{i=1}^N \widehat{r S M I}\left(R^{x_i}, R^{G_\mathcal{XY}\left(x_i\right)}\right)
\end{equation}
where $rSMI$ is the relative squared loss mutual information, $N$ is the number of samples and $R^{(\cdot)}$ denotes the random variables for pixels in $x_i$ and $\mathcal{T}(y)$. Methods like NEGCUT~\cite{wang2021instance} trained a separate generator to generate negative samples dynamically, effectively bringing positive and query examples closer together, whereas F-LeSim~\cite{zheng2021spatially} focused on preserving scene structures by minimizing losses based on spatially-correlative maps. 

\textbf{Medical imaging} Cross modality image synthesis was proposed in~\cite{yu2019ea, yu2020sample} to improve sample quality and efficiency in MR images based on cycle consistency. Similarly, cycle consistency was used for endoscopic image synthesis~\cite{sharan2021mutually} while contrastive learning has been used for medical image segmentation in~\cite{chaitanya2020contrastive, hu2021semi, peng2022boundary}. In this study we focus on developing semantic consistent unpaired image translation of surgical images that differ in modality to MR images and the application proposed so far. Pfeiffer \etal~\cite{pfeiffer2019generating} proposed a variant of MUNIT~\cite{huang2018multimodal} combining MS-SSim~\cite{wang2003multiscale} as a loss for laparoscopic surgery application. This model still falls behind in maintaining semantic consistency (Table~\ref{tab:main}). Hereafter, we refer to this model as \emph{LapMUNIT}. 

\section{Model setup}
Our goal is to maintain the content and semantic correlation between the anatomical structures during translation. In this section, we define various components of the translation approach.

\subsection{Adversarial learning}
GANs~\cite{goodfellow2020generative} have been promising candidates for image translation tasks. The main goal of such an image translation technique is to acquire the ability to map between two domains, $\mathcal{X}$ and $\mathcal{Y}$, based on training samples ${x_i}$ and ${y_j}$ drawn from the distributions $p(X)$ and $p(Y)$ respectively. The generator $G_{\mathcal{XY}}$ learns the mapping between domains and generates the translated image $\mathcal{T}(y)$ and the discriminator $D_{\mathcal{Y}}$ is trained to distinguish between original images $x$ and translated images. The adversarial loss is defined as,  
\begin{equation}
\begin{aligned}
\mathcal{L}_{G A N}\left(G_{\mathcal{XY}}, D_{\mathcal{Y}}\right) & =\mathbb{E}_{y \sim p(Y)}\left[\log D_{\mathcal{Y}}(y)\right] \\
& +\mathbb{E}_{x \sim p(X)}\left[\log \left(1-D_{\mathcal{Y}}\left(G_{\mathcal{XY}}(x)\right)\right) .\right.
\end{aligned}
\end{equation}
Typically, the loss is used to encourage the distributional match between the translated images and images from domain $\mathcal{Y}$. 

\subsection{Patch constrastive learning}
This framework was formulated on noise contrastive estimation (NCE), aiming to maximize the mutual information between the domains. The InfoNCE loss~\cite{oord2018representation} was used to learn 
embeddings between the domains and establish associations between corresponding patches of input and output images while disassociating them if unrelated. 
Let $s$ be the query vector and $s^{+}$ and $s^{-}$ be the positive and negative vectors from the images, respectively. The $s^{-}$ vectors are sampled at $N$ different locations in the input. Finally, the loss is formulated as an (N+1)- way classification and defined as 
\begin{equation}
    \mathcal{L}_{NCE} =  
   -\log \left[\frac{\exp \left(\boldsymbol{s} \cdot \boldsymbol{s}^{+} / \tau\right)}{\exp \left(\boldsymbol{s} \cdot \boldsymbol{s}^{+} / \tau\right)+\sum_{n=1}^N \exp \left(\boldsymbol{s} \cdot \boldsymbol{s}_n^{-} / \tau\right)}\right]
\end{equation}
where $\tau$ is a scaling parameter to factor the distances between the vectors. 

A multilayer patch-based contrastive loss was further employed within the CUT framework, formally defined as \emph{PatchNCE}. 
It leverages the ready availability of the generator $G_{\mathcal{XY}}$ to extract features from $L$ layers at $S$ spatial locations. The \emph{PatchNCE} loss is defined as, 
\begin{equation}
    \mathcal{L}_{\mathrm{Patch}}(X)=\mathbb{E}_{\boldsymbol{x} \sim X} \sum_{l=1}^L \sum_{s=1}^{S} \mathcal{L}_{NCE}
\end{equation}
Despite its aim to promote semantic consistency between input and output images, CUT still faces challenges when the two domains have different semantic characteristics (Section~\ref{sec:results}). This challenge stems from the limited ability of the contrastively learned semantics to enforce correspondence across different domains effectively. 

\subsection{Semantic consistency}
%

Next, we define the \emph{multi-scale structural similarity} (MS-SSim)~\cite{wang2003multiscale} metric. This measure was proposed as a metric for image quality assessment. The extracted structure information from the images is compared on varying image resolutions with a weighting factor for each. 
Initially, given two images, $\mathbf{x}$ and $\mathbf{y}$, let $v_1 = 2 \sigma_{xy} + C_2$ and $v_2 = \sigma_{x}^{2} + \sigma_{y}^{2} + C_2$. Then contrast sensitivity($\mathbf{cs}$) and structure map ($\mathbf{ss}$) are defined as,
\begin{equation} \label{eq:map}
    \operatorname{cs}(\mathbf{x},\mathbf{y}) = \frac{v_1}{v_2}, \quad
    \operatorname{ss}(\mathbf{x},\mathbf{y}) = \frac{(2 \mu_{x} \mu_{y} + C_1) v_1}{(\mu_{x}^{2} + \mu_{y}^{2} + C_1) v_2}
\end{equation}
where $\mu_{(\cdot)}$ and $\sigma_{(\cdot)}$ are the mean and variance and $\sigma_{x,y}$ is the covariance between $\mathbf{x}$ and $\mathbf{y}$. $C_1$ and $C_2$ are small constants depending on the pixel values. The MS-SSim metric is defined as,
\begin{equation}
\operatorname{MS-SSim}(\mathbf{x}, \mathbf{y})=\left[W_i\right]\cdot \prod_{i=1}^K\mathbf{cs}_i \cdot \mathbf{ss}_i
\end{equation}
where $i=1\cdots K$ denotes the number of different image scales and $W_i$ the weight for the $i^{th}$ scale. Hereafter, we mention the loss as \emph{semantic loss}. It is defined as,
\begin{equation}
    \mathcal{L}_{semantic} = 1 - \operatorname{MS-SSim}(x,y)
\end{equation}

\begin{figure}
\begin{center}
\includegraphics[width=0.4\textwidth, height=4cm]{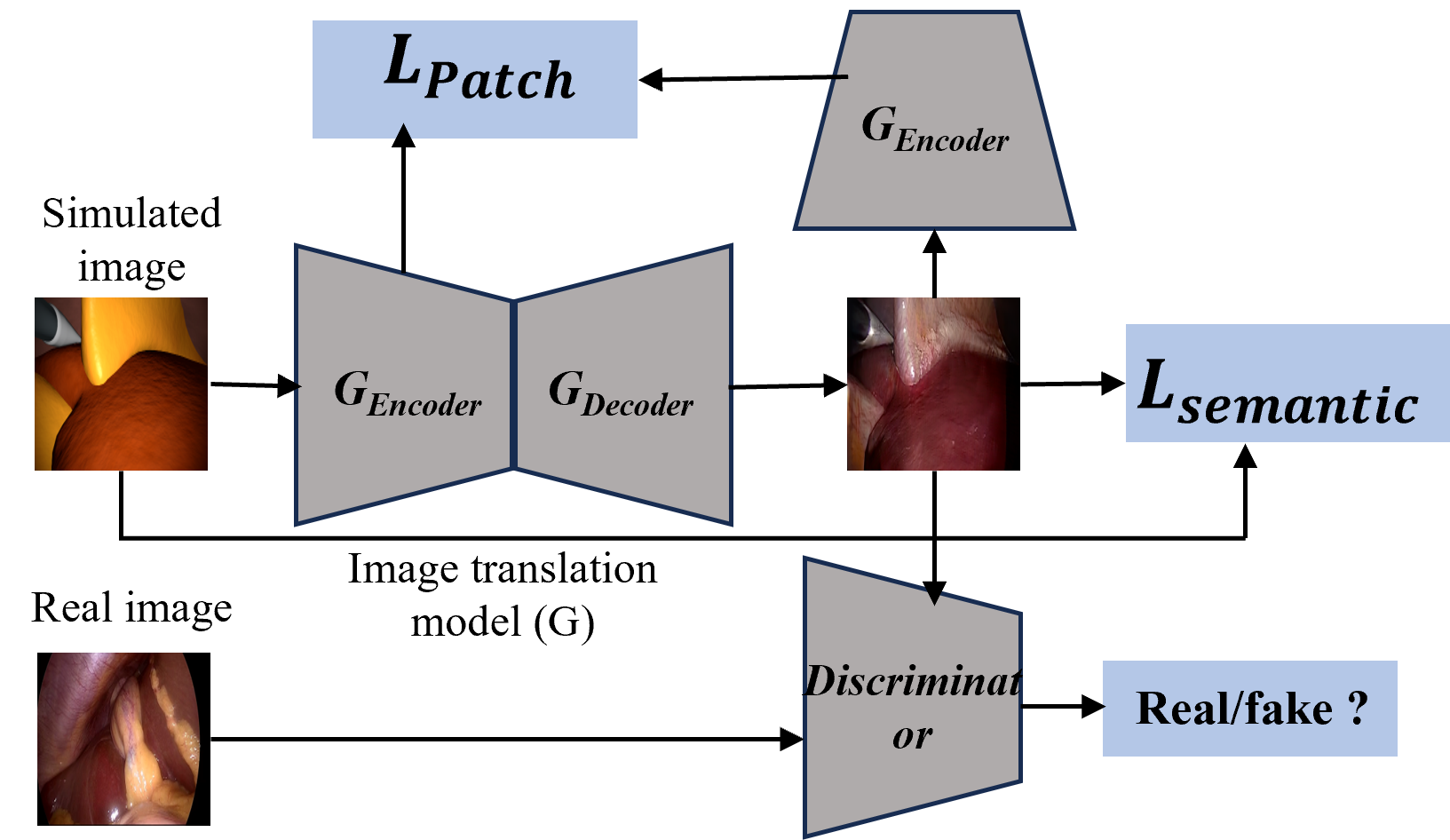}
\end{center}
   \caption{The overview of the ConStructS model with different loss functions.}
\label{fig:model}
\end{figure}




\paragraph{Constrastive learning coupled with MS-SSIM}
The semantic loss concentrates on maintaining the feature structure and considers the lighting conditions. We coupled Contrastive Learning with Structural Similarity (ConStructS) as a model to tackle semantic distortion. To the best of our knowledge, this combination has not been proposed yet. The model overview is shown in Figure~\ref{fig:model}. The generated image should preserve the content information from domain $X$, while the style should be drawn from domain $Y$. Additionally, semantic consistency should to be maintained. The final objective is defined as, 
\begin{equation}
\begin{aligned}
    \mathcal{L}_{total} = \mathcal{L}_{GAN} + \lambda_{x} \mathcal{L}_{Patch}(X) + \lambda_{y}  \mathcal{L}_{Patch}(Y) \\ + \lambda_{ss} \mathcal{L}_{semantic}
\end{aligned}
\end{equation}
where $\lambda_x$, $\lambda_y$, and $\lambda_{ss}$ are weighting parameters for the PatchNCE and semantic losses, respectively. The $\mathcal{L}_{Patch}(Y)$ resembles the identity loss~\cite{zhu2017unpaired} and is applied to the images $y$ to prevent degenerate cases.
\section{Experiments}
In this section, we outline our experiments where the performance of several popular unpaired image translation models namely, CycleGAN~\cite{zhu2017unpaired}, the VGG-based perpetual loss~\cite{johnson2016perceptual}, GcGAN~\cite{fu2019geometry}, DistanceGAN~\cite{tran2018dist},  DRIT${++}$~\cite{lee2018diverse}, LapMUNIT~\cite{pfeiffer2019generating}, UGAT-IT~\cite{kim2019u}, NEGCUT~\cite{wang2021instance} are compared. We demonstrate the effectiveness of \emph{Constructs} in translating synthetic data to the realistic domain with minimal semantic distortion. 
Also, various configurations of contrastive-based models were investigated. The CUT model was trained with the SCC loss~\cite{guo2022alleviating} and SRC~\cite{jung2022exploring}. FeSim and LeSim~\cite{zheng2021spatially} were trained with the CUT as the backbone. The feature perturbated version of CUT, SRUNIT~\cite{jia2021semantically} was also compared.

In particular, the existing baselines exhibit distinct strengths and weaknesses. While certain baselines excel in specific tasks, they may falter in others. Except for LapMUNIT~\cite{pfeiffer2019generating}, no tailored approach exists for surgical scenarios. Consequently, we evaluate ConStructS against several other methods to align with the prevailing research.

Finally, we provide a rationale for the design choices made in the ConStructS model to ensure semantic consistency with an ablation study. We train the model without the semantic loss, which reverts to the basic CUT model~\cite{park2020contrastive} and without the \emph{PatchNCE} loss. Similarly, we combined the semantic loss with cycle consistency into the CycleGAN model for a different combination.

\subsection{Data}
We evaluated the translation methods mentioned above on two different surgical datasets.
\paragraph{Cholecystectomy dataset.} This surgery serves to remove the gallbladder.
For the simulated domain $\mathcal{X}$, we utilized the publicly available synthetic dataset resembling laparoscopic scenes~\cite{pfeiffer2019generating}.
The dataset consists of different anatomical structures such as the liver, liver ligament, gallbladder, abdominal wall, and fat, as well as surgical tools. A total of $20,000$ rendered images forms the synthetic dataset. 
The real images for the domain $\mathcal{Y}$ are taken from 80 videos of the
Cholec80 data set (videos of 80 laparoscopic cholecystectomies)~\cite{twinanda2016endonet}. The videos in which the gallbladder was still intact were identified, and frames were extracted. We finally created a training dataset of approximately $26,000$ images from $75$ patients. A separate segmentation dataset of $5$ patients was chosen. The liver was manually segmented in $196$ images for the downstream evaluation (Section~\ref{eval2}). The images were cropped to $256$ x $512$ pixels, and the training set consists of $17,500$ images, with the remaining $2500$ serving as the test set.

\paragraph{Gastrectomy dataset.}
For this case, we utilized the real and synthetic dataset from~\cite{yoon2022surgical}, based on $40$ real surgical videos of distal gastrectomy.
Along with the surgical tools, five different structures exist the gallbladder, liver, pancreas, spleen, and stomach. The dataset consists of $3400$ synthetic 
and $4500$ real images with corresponding segmentation masks. $2400$ images constituted the training set, with $1000$ images as the test set. The images were resized and cropped to $512$ x $512$ pixels.


\subsection{Evaluation}
The careful selection of appropriate metrics to quantitatively evaluate translation performance is paramount. Our specific focus lies in maintaining semantic consistency and realism during translation. While widely used metrics like FID, KID, and MMD~\cite{heusel2017gans, gretton2012kernel, binkowski2018demystifying} have gained popularity, they do not account for the unmatched semantics inherent in unpaired image translation datasets~\cite{jia2021semantically}. Therefore, we opted for two different quantitative evaluation schemes to overcome these limitations. We also provide qualitative evaluations.

\paragraph{Train:Real$\xrightarrow{}$Eval:Synthetic}Firstly, we adopted the practice of computing metrics based on an \emph{off-the-shelf} segmentation model following~\cite{chu2017cyclegan, fu2019geometry, park2020contrastive, yoon2022surgical}. We train a segmentation model on the real images of the specific dataset. Then the translated synthetic images are tested using this pre-trained model i.e, the metrics are computed against ground truth labels of the synthetic images.  The underlying intuition of this approach is that, if the translation model is able to reduce the domain gap between the real and synthetic images, then the segmentation accuracy from this pre-trained model on the translated synthetic images would be higher~\cite{isola2017image}. This method assesses both the quality, as well as semantic consistency of the translated images. We refer to this method as \emph{eval-$1$}.


\paragraph{Translated images as training data:}\label{eval2}Furthermore, we assess the practical utility of the translated images in a downstream task, as explored in~\cite{pfeiffer2019generating}. We show the usefulness of the translated synthetic images in two different methods. Firstly, we train a segmentation model using only the translated images and evaluate the performance of this model on segmenting the liver organ on a patient-specific categorized dataset consisting of real images. This approach aligns with the intuition mentioned above and provides insights into the realism of the translated images. Secondly, we fine-tune this model on the real data and evaluate them on the same test set of real images. The performance is also compared to a baseline model trained only on real images. By adopting these methodologies, we effectively demonstrate the value of utilizing the translation approach to generate realistic training data. Hereafter, we refer to this method as \emph{eval-$2$}.

\subsection{Implementation details}
We intentionally matched the architectures and hyperparameters to have a reasonable performance comparison to the CUT~\cite{park2020contrastive} and CycleGAN~\cite{zhu2017unpaired} models and their variants. We used a ResNet~\cite{johnson2016perceptual} based generator and a PatchGAN~\cite{isola2017image} as the discriminator—the LS-GAN loss~\cite{mao2017least} used while training the generator with Adam optimizer and a batch size of $1$.
The $\lambda_{x}$ and $\lambda_{y}$ were maintained at a value of $1$. 
Following~\cite{pfeiffer2019generating}, the semantic loss was operated on the images' brightness (average over the channels) as this retains the brightness variations while avoiding the penalization of style-related changes in hue. All the models were trained based on author's code. The same image size was maintained throughout for training of all the baseline models. 


We used a DeepLabV3$+$~\cite{chen2018encoder} model for both evaluation schemes. This model generally performs well for semantic segmentation of surgical images~\cite{rodrigues2022surgical, yoon2022surgical, kolbinger2023anatomy, silva2022analysis}. A $3$-fold cross-validation was performed for \emph{eval-$1$}. We report mean-pixel accuracy (pxAcc), class accuracy (class-weighted pixel accuracy) (clsAcc), and mean intersection-over-union (mIOU) metrics. For the segmentation of the liver on the real images (\emph{eval-$2$}), the median dice scores are reported. For more details on training, the readers can refer to the supplementary material.



\subsection{Results}\label{sec:results}


\begin{table}[t]
  \begin{center}
    {\small{
    \resizebox{\linewidth}{!}{
\begin{tabular}{llccc}
\toprule
Approach & Method & pxAcc & clsAcc & mIOU \\
\midrule
\multirow{7}{2cm}{\emph{Cycle consistency}} & CycleGAN~\cite{zhu2017unpaired} & $0.49\pm 0.08$ & $0.41\pm 0.14$ & $0.23\pm0.09$ \\
& CycleGAN+VGG~\cite{johnson2016perceptual} & $0.52\pm0.09$ & $0.43\pm0.11$ & $0.25\pm0.10$\\
& DRITT$++$~\cite{lee2018diverse_new} & $0.42\pm 0.03$ & $0.28\pm 0.05$ & $0.17\pm 0.04$\\
& LapMUNIT~\cite{pfeiffer2019generating} & $\underline{0.53\pm 0.06}$ & $0.38\pm 0.08$ & $0.25\pm 0.06$\\
& UGAT-IT~\cite{kim2019u} & $0.40\pm 0.03$ & $0.28\pm 0.05$ & $0.16\pm 0.04$ \\\hdashline
\multirow{2}{2cm}{\emph{One-sided translation}}& GcGAN~\cite{fu2019geometry}  & $0.51\pm 0.08$ & $\mathbf{0.44\pm 0.10}$ & $0.26\pm 0.08$\\
& DistGAN~\cite{tran2018dist} & $0.40\pm 0.03$ & $0.28\pm 0.5$ & $0.16\pm 0.04$\\ \hdashline
\multirow{4}{2cm}{\emph{Contrastive learning}}& SRC~\cite{jung2022exploring} & $0.51\pm 0.07$ & $\underline{0.43\pm 0.16}$ & $0.25\pm 0.09$\\
& NEGCUT~\cite{wang2021instance} & $0.49\pm 0.08$ & $0.41\pm 0.15$ & $0.23\pm 0.09$\\
& FeSim~\cite{zheng2021spatially} & $0.41\pm0.10$ & $0.37\pm0.16$ & $0.20\pm0.09$\\
& LeSim~\cite{zheng2021spatially} & $0.47\pm0.09$ & $0.43\pm0.13$ & $0.24\pm0.09$ \\ \hdashline
\multirow{3}{2cm}{\emph{Semantic consistency}} & CycleGAN$+$SCC~\cite{guo2022alleviating}  & $0.50\pm 0.10$ & $0.43\pm 0.15$ & $0.25\pm 0.10$\\ 
& CUT$+$SCC~\cite{guo2022alleviating} & $0.42\pm 0.06$ & $0.35\pm 0.12$ & $0.18\pm 0.07$ \\
& SRUNIT~\cite{jia2021semantically} & $0.50\pm 0.08$ & $0.40\pm0.13$ & $0.23\pm0.08$ \\ \hdashline
\multirow{3}{2cm}{\emph{Ablation study}} & ConStructS w/o $\mathcal{L}_{semantic}$~\cite{park2020contrastive} & $0.50\pm0.07$ & $0.40\pm0.14$ & $\underline{0.26\pm0.09}$\\
& ConStructS w/o PatchNCE & $0.50\pm0.10$ & $0.40\pm0.14$ & $0.25\pm0.10$\\
& CycleGAN$+$ $\mathcal{L}_{semantic}$ & $0.49\pm0.10$ & $0.43\pm0.15$ & $0.24\pm0.09$\\ \hdashline
& ConStructS & $\mathbf{0.59\pm0.07}$ & $\mathbf{0.44\pm0.12}$ & $\mathbf{0.29\pm0.09}$\\
\bottomrule
\end{tabular}
}
}}
\end{center}
\caption{The results of various translation models on the cholecystectomy dataset. \emph{pxAcc} and \emph{clsAcc} denotes the pixel and mean class accuracy respectivly. \emph{mIOU} is the mean intersection over union scores. The best result is indicated in \textbf{bold}, and the second best is \underline{underlined}.}
\label{tab:main}
\end{table}

\begin{figure*}
\begin{center}
\includegraphics[width=\textwidth]{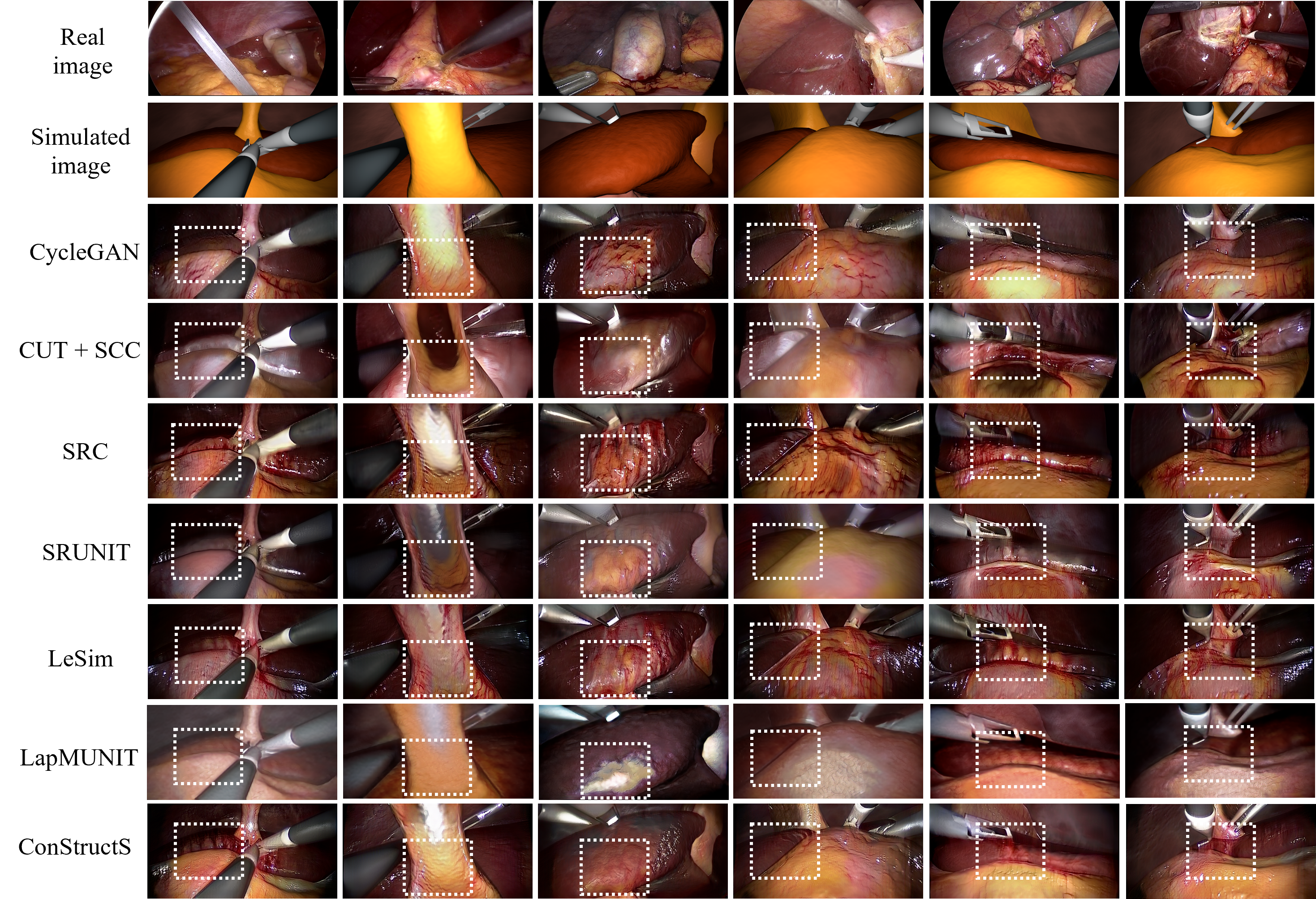}
\end{center}
   \caption{Qualitative results of various translation methods on the cholecystectomy dataset. At the junction of two structures, the textures were interchanged in most of the models. Although not solved completely, the ConStructS model reduces semantic inconsistency. Some regions are highlighted in white boxes.}
\label{fig:main_compare}
\end{figure*}

\paragraph{Cholecystectomy dataset.} The quantitative results are presented in Table~\ref{tab:main}, highlighting the performance of different models. Comparatively, the CycleGAN model with the VGG loss demonstrates better performance than the SCC loss variant. The geometric consistency in GcGAN~\cite{fu2019geometry} leads to a comparable class-accuracy value with ConStructS while outperforming DistGAN~\cite{tran2018dist} and DRIT++~\cite{lee2018diverse_new}. The LapMUNIT~\cite{pfeiffer2019generating} model achieves better scores than the attention-based models. As for the variants of CUT, the addition of SCC loss did not improve its performance further. The SRC~\cite{jung2022exploring} loss helps achieve a class-accuracy score of $0.43$ which coincides with LeSim~\cite{zheng2021spatially}. SRUNIT~\cite{jia2021semantically} and NEGCUT~\cite{wang2021instance} indicate similar performance. Overall, as evidenced by the results, the ConStructS model minimizes semantic distortion to a greater extent and outperforms the recent methods~\cite{jia2021semantically, guo2022alleviating}.

Table~\ref{tab:eval2} indicates the results of \emph{eval-2} method. When the translated images are solely used as training data, the ConStructS model yields comparable results on segmenting the liver to GcGAN~\cite{fu2019geometry} and CycleGAN~\cite{zhu2017unpaired}. A gain of $9\%$ in dice score is obtained compared to the baseline model. Fine-tuning the same model on real data shows that the ConStructS method outperforms the baseline models, showing a $4\%-6\%$ improvement in dice scores. Overall, a $25\%$ improvement is obtained with this model. The qualitative results in Figure~\ref{fig:main_compare} indicate that the ConStructS model reduces the semantic distortion, although not completely, but better than most other translation methods.


\begin{table}
  \begin{center}
    {\small{
    \resizebox{0.85\linewidth}{!}{
\begin{tabular}{lccccc}
\toprule
Method & Pt.76 & Pt.77 & Pt.78 & Pt.79 & Pt.80\\
\midrule
Baseline & $0.69$ & $0.62$ & $0.41$ & $0.60$ & $0.47$ \\
\midrule
\multicolumn{6}{c}{\emph{Train on translated images}} \\
\midrule
CycleGAN~\cite{zhu2017unpaired} & $0.62$ & $0.55$ & $\mathbf{0.67}$ & $0.72$ & $\underline{0.81}$\\
GcGAN~\cite{fu2019geometry} & $0.48$ & $0.52$ & $\underline{0.57}$ & $\mathbf{0.77}$ & $0.79$\\
LapMUNIT~\cite{pfeiffer2019generating} & $0.66$ & $0.64$ & $0.22$ & $0.67$ & $0.60$\\
CUT~\cite{park2020contrastive} & $0.49$ & $0.67$ & $0.40$ & $0.65$ & $0.76$\\
SRUNIT~\cite{jia2021semantically} & $0.64$ & $0.47$ & $0.49$ & $0.68$ & $0.79$\\
CUT$+$SCC~\cite{guo2022alleviating} & $0.44$ & $0.46$ & $0.36$ & $0.49$ & $0.15$\\
SRC~\cite{jung2022exploring} & $0.70$ & $0.67$ & $0.40$ & $0.65$ & $0.76$\\
ConStructS & $\mathbf{0.70}$ & $\mathbf{0.73}$ & $0.28$ & $\underline{0.72}$ & $\mathbf{0.82}$\\
\midrule
\multicolumn{6}{c}{\emph{Pre-train on translated images + fine tune on real images}} \\
\midrule
CycleGAN~\cite{zhu2017unpaired} & $0.80$ & $0.56$ & $0.71$ & $0.81$ & $\underline{0.82}$\\
GcGAN~\cite{fu2019geometry} & $\underline{0.84}$ & $\underline{0.76}$ & $0.71$ & $\underline{0.83}$ & $0.80$\\
LapMUNIT~\cite{pfeiffer2019generating} & $0.80$ & $0.64$ & $0.42$ & $0.76$ & $0.80$\\
CUT~\cite{park2020contrastive} & $0.81$ & $0.63$ & $0.58$ & $0.81$ & $0.75$\\
SRUNIT~\cite{jia2021semantically} & $0.81$ & $0.64$ & $\mathbf{0.75}$ & $0.76$ & $0.81$\\
CUT$+$SCC~\cite{guo2022alleviating} & $0.79$ & $0.56$ & $0.69$ & $0.74$ & $0.40$\\
SRC~\cite{jung2022exploring} & $0.74$ & $0.64$ & $0.64$ & $0.80$ & $0.82$\\
ConStructS & $\mathbf{0.88}$ & $\mathbf{0.82}$ & $\underline{0.74}$ & $\mathbf{0.87}$ & $\mathbf{0.86}$\\
\bottomrule
\end{tabular}
}
}}
\end{center}
\caption{The quantitative results for eval-2 method. Pt. refers to a patient, and the dice scores are reported. The baseline model is only trained on real images. When training with translated images from ConStructS and fine-tuning on real images leads to large gain in segmentation performance. }
\label{tab:eval2}
\end{table}

\begin{figure}
\begin{center}
\includegraphics[width=0.4\textwidth]{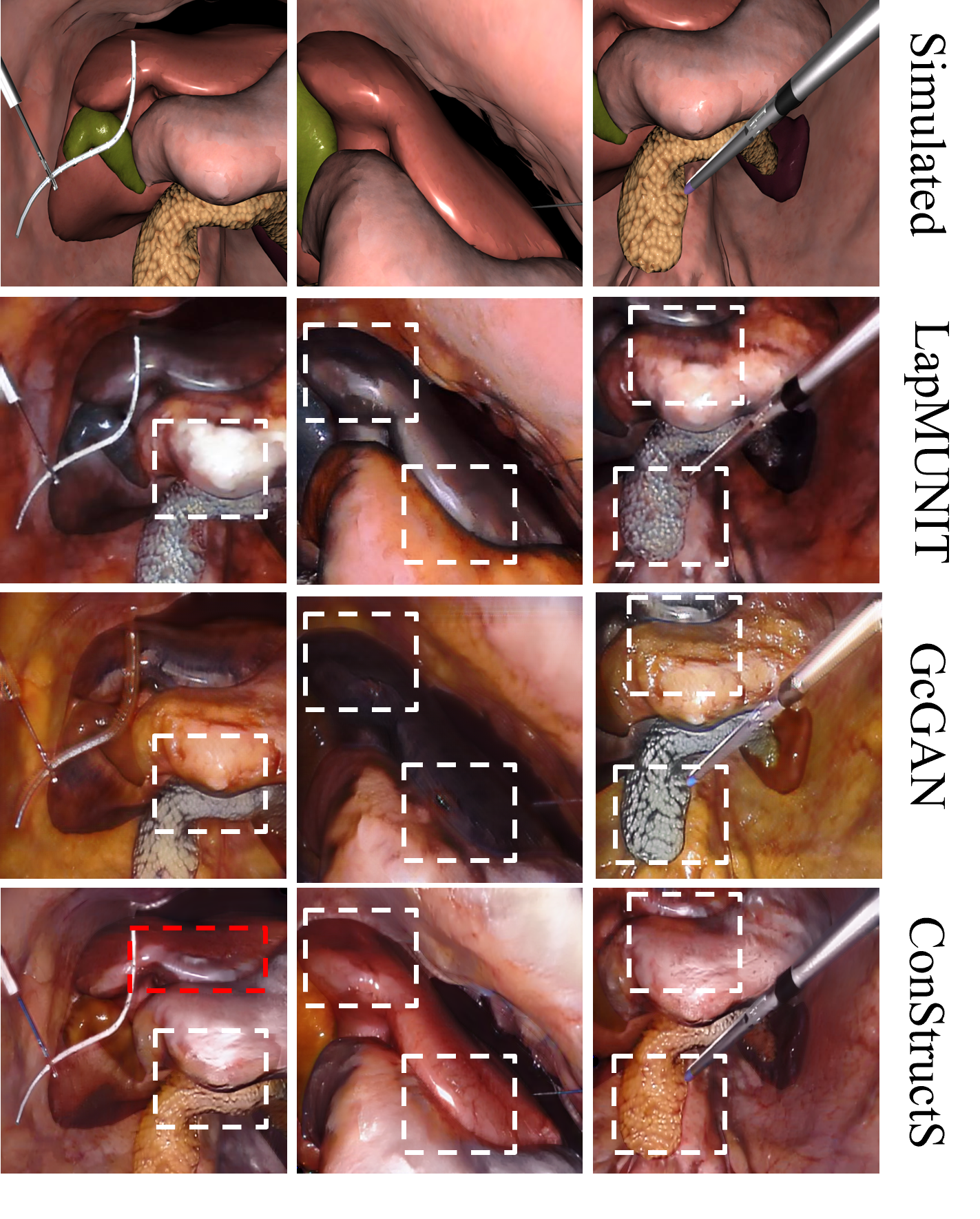}
\end{center}
   \caption{Qualitative samples from the gastrectomy dataset. The white boxes highlight some regions. The red box indicates one of the failure cases of ConStructS, where a tool-like texture is mapped on the liver.}
\label{fig:gast}
\end{figure}

\paragraph{Gastrectomy dataset.} The visual results depicted in Figure~\ref{fig:gast} demonstrate that ConStructS significantly mitigates semantic mismatches, particularly in regions characterized by differing specularity compared to other models. As presented in Table~\ref{tab:gast}, quantitative analysis reveals that LapMUNIT~\cite{pfeiffer2019generating} outperforms both GcGAN~\cite{fu2019geometry} and CycleGAN~\cite{zhu2017unpaired} models. Conversely, the ConStructS model exhibits a moderate improvement in performance compared to all the other models.

\begin{table}
  \begin{center}
    {\small{
    \resizebox{\linewidth}{!}{
\begin{tabular}{lccc}
\toprule
Method & pxAcc & clsACC & mIOU\\
\midrule
CycleGAN~\cite{zhu2017unpaired} & $0.39\pm0.12$ & $0.17\pm0.14$ & $0.09\pm0.10$\\
GcGAN~\cite{fu2019geometry} & $0.40\pm0.13$ & $0.18\pm0.01$ & $0.10\pm0.01$\\
LapMUNIT~\cite{pfeiffer2019generating} & $0.43\pm0.01$ & $0.21\pm0.10$ & $0.11\pm0.09$\\
CUT~\cite{park2020contrastive} & $0.42\pm0.01$ & $\underline{0.22\pm0.02}$ & $\mathbf{0.11\pm0.05}$\\
SRUNIT~\cite{jia2021semantically} & $\underline{0.44\pm0.01}$ & $0.20\pm0.01$ & $0.10\pm0.05$\\
SRC~\cite{jung2022exploring} & $0.44\pm0.02$ & $0.19\pm0.15$ & $0.09\pm0.10$\\
ConStructS & $\mathbf{0.46\pm0.08}$ & $\mathbf{0.24\pm0.13}$ & $\underline{0.10\pm0.09}$\\
\bottomrule
\end{tabular}
}
}}
\end{center}
\caption{The quantitative results of the translational models on the gastrectomy dataset.}
\label{tab:gast}
\end{table}

\begin{figure*}
\begin{center}
\includegraphics[width=0.95\textwidth]{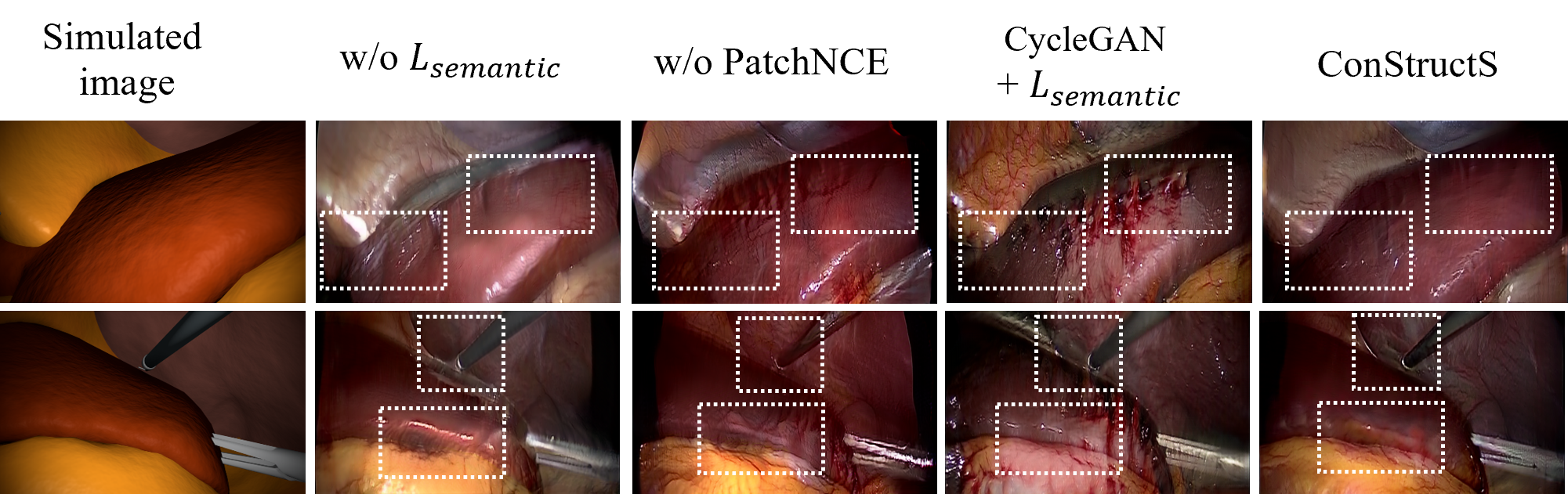}
\end{center}
   \caption{Qualitative results of the ablation study on the cholecystectomy dataset. Texture mismatch occurs in low-lighting regions without the semantic loss. As seen from the $2^{nd}$ row without the PatchNCE loss, no explicit boundary exists between the liver and abdominal wall leading to both regions having the same semantic textures.}
\label{fig:ablation}
\end{figure*}

\begin{figure}
\begin{center}
\includegraphics[width=0.48\textwidth, height=2cm]{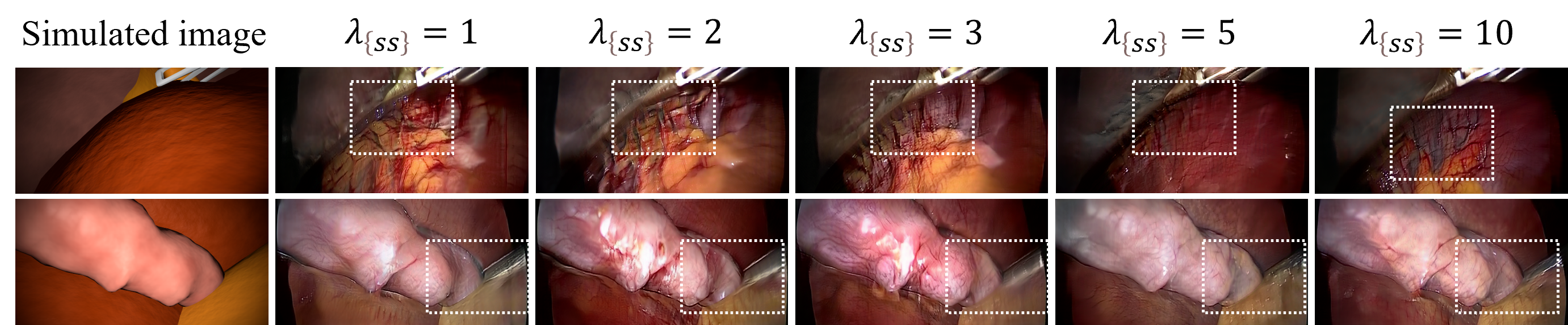}
\end{center}
   \caption{Sensitivity analysis examples on the cholecystectomy dataset. For the $\lambda_{SS}$ value being $5$, liver texture is maintained ($1^{st}$ row) and the tool texture (grey lining) is avoided between the junction of the structures ($2^{nd}$ row).}
\label{fig:sens}
\end{figure}

\paragraph{Ablation study.} The qualitative results of the ablation study are presented in Figure~\ref{fig:ablation}. When examining the CUT model, specifically ConStructS, without semantic loss, we observe that the structure is well preserved during translation. However, there is a noticeable mismatch in texture in regions with reduced brightness, which can be attributed to variations in lighting conditions. In the absence of the PatchNCE loss, as there is no explicit control over image patches, structure information is mixed, resulting in the mapping of styles from different structures (e.g., fat or blood) to unlikely regions. Lastly, the combination of the semantic loss with the CycleGAN model yields an improvement compared to the basic CycleGAN model. Regardless, as seen from Table~\ref{tab:main}, this combination still lacks performance. 
\paragraph{Sensitivity analysis.}
In this section, we study the sensitivity of the parameter $\lambda_{SS}$ and the direct influence of the \emph{semantic} loss. The parameter $\lambda_{SS}$ is varied between the values $1,2,3,5$ and $10$. As the $\lambda_{SS}$ value is increased, there is a performance improvement up to a certain threshold. Figure~\ref{fig:sens} and Table~\ref{tab:sens} indicate that setting the appropriate $\lambda_{SS}$ (here, $5$) effectively controls the irregular texture between the gallbladder and the tool. However, it is necessary to note that large values limit the benefits of the semantic loss, as the model primarily focuses on reducing structure distortion and disregards the style information. A binary search could identify the largest $\lambda_{SS}$ value that maintains the semantic character.

\begin{table}
  \begin{center}
    {\small{
\begin{tabular}{lccc}
\toprule
$\lambda_{SS}$ & pxAcc & clsACC & mIOU\\
\midrule
$1$ & $0.50\pm0.06$ & $0.42\pm0.14$ & $0.24\pm0.08$\\
$2$ & $0.52\pm0.07$ & $0.43\pm0.14$ & $0.25\pm0.09$\\
$3$ & $0.54\pm0.06$ & $0.43\pm0.13$ & $0.25\pm0.08$\\
$5$ & $0.59\pm0.07$ & $0.44\pm0.12$ & $0.29\pm0.09$\\
$10$ & $0.56\pm0.08$ & $0.42\pm0.13$ & $0.27\pm0.09$\\
\bottomrule
\end{tabular}
}}
\end{center}
\caption{Quantitative results of the sensitivity analysis on $\lambda_{SS}$ for the cholecystectomy dataset.}
\label{tab:sens}
\end{table}

\subsection{Discussion}

Traditional approaches prioritizing distance preservation, such as DistanceGAN~\cite{tran2018dist} or the $L_1$ reconstruction loss used in CycleGAN~\cite{zhu2017unpaired}, typically do not effectively enhance semantic consistency. These pixel-based metrics tend to be highly sensitive to structural transformations and variations in lighting conditions, which can introduce artifacts in the generated images during translation. While SRUNIT~\cite{jia2021semantically} shows promise in reducing semantic distortion by introducing perturbations in the feature space, it alone is insufficient for the specific application at hand. On the contrary, the NEGCUT~\cite{wang2021instance} model aims to preserve the overall structure during translation but needs to be more accurate in mapping textures between these structures. The same limitation has been observed in the LeSim~\cite{zheng2021spatially} model.

The results of our ablation study demonstrate the crucial role of combining PatchNCE with semantic loss in mitigating semantic distortion. We posit that leveraging the contrastive learning approach on the generator's encoded feature space makes learning higher-level attributes, such as organ or tool structures, possible. However, relying solely on this aspect for matching semantic information has limitations~\cite{jia2021semantically}. To address this, we introduced the semantic loss as a regularizer that operates on the multiple scales of the images (i.e., different resolutions), checking for the structure consistency and lighting conditions (Equation~\ref{eq:map}). The combination of PatchNCE and semantic loss proves effective in preserving the semantic characteristics throughout the translation process.


\paragraph{Limitations.} The combination of semantic loss and contrastive learning holds promise for mitigating semantic inconsistencies; however, it is essential to acknowledge its limitations and potential failure cases. Achieving a comprehensive and universal solution to the semantic distortion challenge solely through a single image-to-image translation method is a complex task. Existing approaches primarily focusing on one-sided translation have overlooked the synthesis of multi-modal data. In this context, the ConStructS model emerges as a promising candidate for future exploration and research. By incorporating multi-modal outcomes, this model can be utilized to generate diverse data with improved semantic consistency.

\section{Conclusion}
In this study, we conducted an empirical investigation on the issue of semantic inconsistency in unpaired image translation, focusing on its relevance to surgical applications where labeled data is minimal. We extensively evaluate several state-of-the-art unpaired translation methods, explicitly targeting the translation of images from a simulated domain to a realistic environment. Addressing the problem of semantic distortion, we found a novel combination of a structure similarity metric with contrastive learning as the most effective. Surprisingly, this simple model reduces semantic distortion while preserving the realism of the translated images and shows the highest utility as training data for downstream tasks.
\section{Acknowledgements}
This work is partly supported by BMBF (Federal Ministry of Education and Research) in DAAD project 57616814 (\href{https://secai.org/}{SECAI, School of Embedded Composite AI}) as part of the program Konrad Zuse Schools of Excellence in Artificial Intelligence. This work is partly funded by the German Research Foundation (DFG, Deutsche Forschungsgemeinschaft) as part of Germany’s Excellence Strategy – EXC 2050/1 –Project ID 390696704 – Cluster of Excellence “Centre for Tactile Internet with Human-in-the-Loop” (CeTI) of Technische Universit¨at Dresden.

{\small
\bibliographystyle{ieee_fullname}
\bibliography{ref}

\begin{thebibliography}{10}\itemsep=-1pt

\bibitem{alami2018unsupervised}
Youssef Alami~Mejjati, Christian Richardt, James Tompkin, Darren Cosker, and
  Kwang~In Kim.
\newblock Unsupervised attention-guided image-to-image translation.
\newblock {\em Advances in neural information processing systems}, 31, 2018.

\bibitem{amodio2019travelgan}
Matthew Amodio and Smita Krishnaswamy.
\newblock Travelgan: Image-to-image translation by transformation vector
  learning.
\newblock In {\em Proceedings of the ieee/cvf conference on computer vision and
  pattern recognition}, pages 8983--8992, 2019.

\bibitem{benaim2017one}
Sagie Benaim and Lior Wolf.
\newblock One-sided unsupervised domain mapping.
\newblock {\em Advances in neural information processing systems}, 30, 2017.

\bibitem{binkowski2018demystifying}
Miko{\l}aj Bi{\'n}kowski, Danica~J Sutherland, Michael Arbel, and Arthur
  Gretton.
\newblock Demystifying mmd gans.
\newblock {\em arXiv preprint arXiv:1801.01401}, 2018.

\bibitem{bodenstedt2020artificial}
Sebastian Bodenstedt, Martin Wagner, Beat~Peter M{\"u}ller-Stich, J{\"u}rgen
  Weitz, and Stefanie Speidel.
\newblock Artificial intelligence-assisted surgery: potential and challenges.
\newblock {\em Visceral Medicine}, 36(6):450--455, 2020.

\bibitem{chaitanya2020contrastive}
Krishna Chaitanya, Ertunc Erdil, Neerav Karani, and Ender Konukoglu.
\newblock Contrastive learning of global and local features for medical image
  segmentation with limited annotations.
\newblock {\em Advances in neural information processing systems},
  33:12546--12558, 2020.

\bibitem{chen2018encoder}
Liang-Chieh Chen, Yukun Zhu, George Papandreou, Florian Schroff, and Hartwig
  Adam.
\newblock Encoder-decoder with atrous separable convolution for semantic image
  segmentation.
\newblock In {\em Proceedings of the European conference on computer vision
  (ECCV)}, pages 801--818, 2018.

\bibitem{choi2018stargan}
Yunjey Choi, Minje Choi, Munyoung Kim, Jung-Woo Ha, Sunghun Kim, and Jaegul
  Choo.
\newblock Stargan: Unified generative adversarial networks for multi-domain
  image-to-image translation.
\newblock In {\em Proceedings of the IEEE conference on computer vision and
  pattern recognition}, pages 8789--8797, 2018.

\bibitem{chu2017cyclegan}
Casey Chu, Andrey Zhmoginov, and Mark Sandler.
\newblock Cyclegan, a master of steganography.
\newblock {\em arXiv preprint arXiv:1712.02950}, 2017.

\bibitem{cordts2016cityscapes}
Marius Cordts, Mohamed Omran, Sebastian Ramos, Timo Rehfeld, Markus Enzweiler,
  Rodrigo Benenson, Uwe Franke, Stefan Roth, and Bernt Schiele.
\newblock The cityscapes dataset for semantic urban scene understanding.
\newblock In {\em Proceedings of the IEEE conference on computer vision and
  pattern recognition}, pages 3213--3223, 2016.

\bibitem{dosovitskiy2016generating}
Alexey Dosovitskiy and Thomas Brox.
\newblock Generating images with perceptual similarity metrics based on deep
  networks.
\newblock {\em Advances in neural information processing systems}, 29, 2016.

\bibitem{fu2019geometry}
Huan Fu, Mingming Gong, Chaohui Wang, Kayhan Batmanghelich, Kun Zhang, and
  Dacheng Tao.
\newblock Geometry-consistent generative adversarial networks for one-sided
  unsupervised domain mapping.
\newblock In {\em Proceedings of the IEEE/CVF Conference on Computer Vision and
  Pattern Recognition}, pages 2427--2436, 2019.

\bibitem{goodfellow2020generative}
Ian Goodfellow, Jean Pouget-Abadie, Mehdi Mirza, Bing Xu, David Warde-Farley,
  Sherjil Ozair, Aaron Courville, and Yoshua Bengio.
\newblock Generative adversarial networks.
\newblock {\em Communications of the ACM}, 63(11):139--144, 2020.

\bibitem{gretton2012kernel}
Arthur Gretton, Karsten~M Borgwardt, Malte~J Rasch, Bernhard Sch{\"o}lkopf, and
  Alexander Smola.
\newblock A kernel two-sample test.
\newblock {\em The Journal of Machine Learning Research}, 13(1):723--773, 2012.

\bibitem{guo2022alleviating}
Jiaxian Guo, Jiachen Li, Huan Fu, Mingming Gong, Kun Zhang, and Dacheng Tao.
\newblock Alleviating semantics distortion in unsupervised low-level
  image-to-image translation via structure consistency constraint.
\newblock In {\em Proceedings of the IEEE/CVF Conference on Computer Vision and
  Pattern Recognition}, pages 18249--18259, 2022.

\bibitem{hager2020surgical}
Gregory~D Hager, Lena Maier-Hein, and S~Swaroop Vedula.
\newblock Surgical data science.
\newblock In {\em Handbook of Medical Image Computing and Computer Assisted
  Intervention}, pages 931--952. Elsevier, 2020.

\bibitem{haidegger2022robot}
Tamas Haidegger, Stefanie Speidel, Danail Stoyanov, and Richard~M Satava.
\newblock Robot-assisted minimally invasive surgery—surgical robotics in the
  data age.
\newblock {\em Proceedings of the IEEE}, 110(7):835--846, 2022.

\bibitem{heusel2017gans}
Martin Heusel, Hubert Ramsauer, Thomas Unterthiner, Bernhard Nessler, and Sepp
  Hochreiter.
\newblock Gans trained by a two time-scale update rule converge to a local nash
  equilibrium.
\newblock {\em Advances in neural information processing systems}, 30, 2017.

\bibitem{hoffman2018cycada}
Judy Hoffman, Eric Tzeng, Taesung Park, Jun-Yan Zhu, Phillip Isola, Kate
  Saenko, Alexei Efros, and Trevor Darrell.
\newblock Cycada: Cycle-consistent adversarial domain adaptation.
\newblock In {\em International conference on machine learning}, pages
  1989--1998. Pmlr, 2018.

\bibitem{hong2020cholecseg8k}
W-Y Hong, C-L Kao, Y-H Kuo, J-R Wang, W-L Chang, and C-S Shih.
\newblock Cholecseg8k: a semantic segmentation dataset for laparoscopic
  cholecystectomy based on cholec80.
\newblock {\em arXiv preprint arXiv:2012.12453}, 2020.

\bibitem{hu2021semi}
Xinrong Hu, Dewen Zeng, Xiaowei Xu, and Yiyu Shi.
\newblock Semi-supervised contrastive learning for label-efficient medical
  image segmentation.
\newblock In {\em Medical Image Computing and Computer Assisted
  Intervention--MICCAI 2021: 24th International Conference, Strasbourg, France,
  September 27--October 1, 2021, Proceedings, Part II 24}, pages 481--490.
  Springer, 2021.

\bibitem{huang2018multimodal}
Xun Huang, Ming-Yu Liu, Serge Belongie, and Jan Kautz.
\newblock Multimodal unsupervised image-to-image translation.
\newblock In {\em Proceedings of the European conference on computer vision
  (ECCV)}, pages 172--189, 2018.

\bibitem{isola2017image}
Phillip Isola, Jun-Yan Zhu, Tinghui Zhou, and Alexei~A Efros.
\newblock Image-to-image translation with conditional adversarial networks.
\newblock In {\em Proceedings of the IEEE conference on computer vision and
  pattern recognition}, pages 1125--1134, 2017.

\bibitem{jia2021semantically}
Zhiwei Jia, Bodi Yuan, Kangkang Wang, Hong Wu, David Clifford, Zhiqiang Yuan,
  and Hao Su.
\newblock Semantically robust unpaired image translation for data with
  unmatched semantics statistics.
\newblock In {\em Proceedings of the IEEE/CVF International Conference on
  Computer Vision}, pages 14273--14283, 2021.

\bibitem{johnson2016perceptual}
Justin Johnson, Alexandre Alahi, and Li Fei-Fei.
\newblock Perceptual losses for real-time style transfer and super-resolution.
\newblock In {\em Computer Vision--ECCV 2016: 14th European Conference,
  Amsterdam, The Netherlands, October 11-14, 2016, Proceedings, Part II 14},
  pages 694--711. Springer, 2016.

\bibitem{jung2022exploring}
Chanyong Jung, Gihyun Kwon, and Jong~Chul Ye.
\newblock Exploring patch-wise semantic relation for contrastive learning in
  image-to-image translation tasks.
\newblock In {\em Proceedings of the IEEE/CVF Conference on Computer Vision and
  Pattern Recognition}, pages 18260--18269, 2022.

\bibitem{kim2019u}
Junho Kim, Minjae Kim, Hyeonwoo Kang, and Kwanghee Lee.
\newblock U-gat-it: Unsupervised generative attentional networks with adaptive
  layer-instance normalization for image-to-image translation.
\newblock {\em arXiv preprint arXiv:1907.10830}, 2019.

\bibitem{kingma2014adam}
Diederik~P Kingma and Jimmy Ba.
\newblock Adam: A method for stochastic optimization.
\newblock {\em arXiv preprint arXiv:1412.6980}, 2014.

\bibitem{kolbinger2023anatomy}
Fiona~R Kolbinger, Franziska~M Rinner, Alexander~C Jenke, Matthias Carstens,
  Stefanie Krell, Stefan Leger, Marius Distler, J{\"u}rgen Weitz, Stefanie
  Speidel, and Sebastian Bodenstedt.
\newblock Anatomy segmentation in laparoscopic surgery: comparison of machine
  learning and human expertise--an experimental study.
\newblock {\em International Journal of Surgery}, pages 10--1097, 2023.

\bibitem{ledig2017photo}
Christian Ledig, Lucas Theis, Ferenc Husz{\'a}r, Jose Caballero, Andrew
  Cunningham, Alejandro Acosta, Andrew Aitken, Alykhan Tejani, Johannes Totz,
  Zehan Wang, et~al.
\newblock Photo-realistic single image super-resolution using a generative
  adversarial network.
\newblock In {\em Proceedings of the IEEE conference on computer vision and
  pattern recognition}, pages 4681--4690, 2017.

\bibitem{lee2018diverse_new}
Hsin-Ying Lee, Hung-Yu Tseng, Jia-Bin Huang, Maneesh Singh, and Ming-Hsuan
  Yang.
\newblock Diverse image-to-image translation via disentangled representations.
\newblock In {\em Proceedings of the European conference on computer vision
  (ECCV)}, pages 35--51, 2018.

\bibitem{lee2018diverse}
Hsin-Ying Lee, Hung-Yu Tseng, Jia-Bin Huang, Maneesh Singh, and Ming-Hsuan
  Yang.
\newblock Diverse image-to-image translation via disentangled
  representations-drit++.
\newblock In {\em Proceedings of the European conference on computer vision
  (ECCV)}, pages 35--51, 2018.

\bibitem{liu2017unsupervised}
Ming-Yu Liu, Thomas Breuel, and Jan Kautz.
\newblock Unsupervised image-to-image translation networks.
\newblock {\em Advances in neural information processing systems}, 30, 2017.

\bibitem{liu2019few}
Ming-Yu Liu, Xun Huang, Arun Mallya, Tero Karras, Timo Aila, Jaakko Lehtinen,
  and Jan Kautz.
\newblock Few-shot unsupervised image-to-image translation.
\newblock In {\em Proceedings of the IEEE/CVF international conference on
  computer vision}, pages 10551--10560, 2019.

\bibitem{maier2022surgical}
Lena Maier-Hein, Matthias Eisenmann, Duygu Sarikaya, Keno M{\"a}rz, Toby
  Collins, Anand Malpani, Johannes Fallert, Hubertus Feussner, Stamatia
  Giannarou, Pietro Mascagni, et~al.
\newblock Surgical data science--from concepts toward clinical translation.
\newblock {\em Medical image analysis}, 76:102306, 2022.

\bibitem{maier2017surgical}
Lena Maier-Hein, Swaroop~S Vedula, Stefanie Speidel, Nassir Navab, Ron Kikinis,
  Adrian Park, Matthias Eisenmann, Hubertus Feussner, Germain Forestier,
  Stamatia Giannarou, et~al.
\newblock Surgical data science for next-generation interventions.
\newblock {\em Nature Biomedical Engineering}, 1(9):691--696, 2017.

\bibitem{mao2017least}
Xudong Mao, Qing Li, Haoran Xie, Raymond~YK Lau, Zhen Wang, and Stephen
  Paul~Smolley.
\newblock Least squares generative adversarial networks.
\newblock In {\em Proceedings of the IEEE international conference on computer
  vision}, pages 2794--2802, 2017.

\bibitem{mechrez2018contextual}
Roey Mechrez, Itamar Talmi, and Lihi Zelnik-Manor.
\newblock The contextual loss for image transformation with non-aligned data.
\newblock In {\em Proceedings of the European conference on computer vision
  (ECCV)}, pages 768--783, 2018.

\bibitem{miyato2018spectral}
Takeru Miyato, Toshiki Kataoka, Masanori Koyama, and Yuichi Yoshida.
\newblock Spectral normalization for generative adversarial networks.
\newblock {\em arXiv preprint arXiv:1802.05957}, 2018.

\bibitem{oord2018representation}
Aaron van~den Oord, Yazhe Li, and Oriol Vinyals.
\newblock Representation learning with contrastive predictive coding.
\newblock {\em arXiv preprint arXiv:1807.03748}, 2018.

\bibitem{park2020contrastive}
Taesung Park, Alexei~A Efros, Richard Zhang, and Jun-Yan Zhu.
\newblock Contrastive learning for unpaired image-to-image translation.
\newblock In {\em Computer Vision--ECCV 2020: 16th European Conference,
  Glasgow, UK, August 23--28, 2020, Proceedings, Part IX 16}, pages 319--345.
  Springer, 2020.

\bibitem{peng2022boundary}
Jizong Peng, Ping Wang, Marco Pedersoli, and Christian Desrosiers.
\newblock Boundary-aware information maximization for self-supervised medical
  image segmentation.
\newblock {\em arXiv preprint arXiv:2202.02371}, 2022.

\bibitem{pfeiffer2019generating}
Micha Pfeiffer, Isabel Funke, Maria~R Robu, Sebastian Bodenstedt, Leon
  Strenger, Sandy Engelhardt, Tobias Ro{\ss}, Matthew~J Clarkson, Kurinchi
  Gurusamy, Brian~R Davidson, et~al.
\newblock Generating large labeled data sets for laparoscopic image processing
  tasks using unpaired image-to-image translation.
\newblock In {\em Medical Image Computing and Computer Assisted
  Intervention--MICCAI 2019: 22nd International Conference, Shenzhen, China,
  October 13--17, 2019, Proceedings, Part V 22}, pages 119--127. Springer,
  2019.

\bibitem{rivoir2021long}
Dominik Rivoir, Micha Pfeiffer, Reuben Docea, Fiona Kolbinger, Carina Riediger,
  J{\"u}rgen Weitz, and Stefanie Speidel.
\newblock Long-term temporally consistent unpaired video translation from
  simulated surgical 3d data.
\newblock In {\em Proceedings of the IEEE/CVF International Conference on
  Computer Vision}, pages 3343--3353, 2021.

\bibitem{rodrigues2022surgical}
Mark Rodrigues, Michael Mayo, and Panos Patros.
\newblock Surgical tool datasets for machine learning research: a survey.
\newblock {\em International Journal of Computer Vision}, 130(9):2222--2248,
  2022.

\bibitem{rombach2022high}
Robin Rombach, Andreas Blattmann, Dominik Lorenz, Patrick Esser, and Bj{\"o}rn
  Ommer.
\newblock High-resolution image synthesis with latent diffusion models.
\newblock In {\em Proceedings of the IEEE/CVF Conference on Computer Vision and
  Pattern Recognition}, pages 10684--10695, 2022.

\bibitem{saharia2022palette}
Chitwan Saharia, William Chan, Huiwen Chang, Chris Lee, Jonathan Ho, Tim
  Salimans, David Fleet, and Mohammad Norouzi.
\newblock Palette: Image-to-image diffusion models.
\newblock In {\em ACM SIGGRAPH 2022 Conference Proceedings}, pages 1--10, 2022.

\bibitem{sharan2021mutually}
Lalith Sharan, Gabriele Romano, Sven Koehler, Halvar Kelm, Matthias Karck,
  Raffaele De~Simone, and Sandy Engelhardt.
\newblock Mutually improved endoscopic image synthesis and landmark detection
  in unpaired image-to-image translation.
\newblock {\em IEEE Journal of Biomedical and Health Informatics},
  26(1):127--138, 2021.

\bibitem{shrivastava2017learning}
Ashish Shrivastava, Tomas Pfister, Oncel Tuzel, Joshua Susskind, Wenda Wang,
  and Russell Webb.
\newblock Learning from simulated and unsupervised images through adversarial
  training.
\newblock In {\em Proceedings of the IEEE conference on computer vision and
  pattern recognition}, pages 2107--2116, 2017.

\bibitem{silva2022analysis}
Bruno Silva, Bruno Oliveira, Pedro Morais, LR Buschle, Jorge Correia-Pinto,
  Estev{\~a}o Lima, and Joao~L Vila{\c{c}}a.
\newblock Analysis of current deep learning networks for semantic segmentation
  of anatomical structures in laparoscopic surgery.
\newblock In {\em 2022 44th Annual International Conference of the IEEE
  Engineering in Medicine \& Biology Society (EMBC)}, pages 3502--3505. IEEE,
  2022.

\bibitem{smith2019super}
Leslie~N Smith and Nicholay Topin.
\newblock Super-convergence: Very fast training of neural networks using large
  learning rates.
\newblock In {\em Artificial intelligence and machine learning for multi-domain
  operations applications}, volume 11006, pages 369--386. SPIE, 2019.

\bibitem{su2022dual}
Xuan Su, Jiaming Song, Chenlin Meng, and Stefano Ermon.
\newblock Dual diffusion implicit bridges for image-to-image translation.
\newblock {\em arXiv preprint arXiv:2203.08382}, 2022.

\bibitem{taigman2016unsupervised}
Yaniv Taigman, Adam Polyak, and Lior Wolf.
\newblock Unsupervised cross-domain image generation.
\newblock {\em arXiv preprint arXiv:1611.02200}, 2016.

\bibitem{tang2021attentiongan}
Hao Tang, Hong Liu, Dan Xu, Philip~HS Torr, and Nicu Sebe.
\newblock Attentiongan: Unpaired image-to-image translation using
  attention-guided generative adversarial networks.
\newblock {\em IEEE transactions on neural networks and learning systems},
  2021.

\bibitem{tran2018dist}
Ngoc-Trung Tran, Tuan-Anh Bui, and Ngai-Man Cheung.
\newblock Dist-gan: An improved gan using distance constraints.
\newblock In {\em Proceedings of the European conference on computer vision
  (ECCV)}, pages 370--385, 2018.

\bibitem{twinanda2016endonet}
Andru~P Twinanda, Sherif Shehata, Didier Mutter, Jacques Marescaux, Michel
  De~Mathelin, and Nicolas Padoy.
\newblock Endonet: a deep architecture for recognition tasks on laparoscopic
  videos.
\newblock {\em IEEE transactions on medical imaging}, 36(1):86--97, 2016.

\bibitem{labelme}
Kentaro Wada.
\newblock {Image Polygonal annotation with Python}.
\newblock \url{https://github.com/labelmeai/labelme}.

\bibitem{wang2022pretraining}
Tengfei Wang, Ting Zhang, Bo Zhang, Hao Ouyang, Dong Chen, Qifeng Chen, and
  Fang Wen.
\newblock Pretraining is all you need for image-to-image translation.
\newblock {\em arXiv preprint arXiv:2205.12952}, 2022.

\bibitem{wang2018high}
Ting-Chun Wang, Ming-Yu Liu, Jun-Yan Zhu, Andrew Tao, Jan Kautz, and Bryan
  Catanzaro.
\newblock High-resolution image synthesis and semantic manipulation with
  conditional gans.
\newblock In {\em Proceedings of the IEEE conference on computer vision and
  pattern recognition}, pages 8798--8807, 2018.

\bibitem{wang2022semantic}
Weilun Wang, Jianmin Bao, Wengang Zhou, Dongdong Chen, Dong Chen, Lu Yuan, and
  Houqiang Li.
\newblock Semantic image synthesis via diffusion models.
\newblock {\em arXiv preprint arXiv:2207.00050}, 2022.

\bibitem{wang2021instance}
Weilun Wang, Wengang Zhou, Jianmin Bao, Dong Chen, and Houqiang Li.
\newblock Instance-wise hard negative example generation for contrastive
  learning in unpaired image-to-image translation.
\newblock In {\em Proceedings of the IEEE/CVF International Conference on
  Computer Vision}, pages 14020--14029, 2021.

\bibitem{wang2003multiscale}
Zhou Wang, Eero~P Simoncelli, and Alan~C Bovik.
\newblock Multiscale structural similarity for image quality assessment.
\newblock In {\em The Thrity-Seventh Asilomar Conference on Signals, Systems \&
  Computers, 2003}, volume~2, pages 1398--1402. Ieee, 2003.

\bibitem{yoon2022surgical}
Jihun Yoon, SeulGi Hong, Seungbum Hong, Jiwon Lee, Soyeon Shin, Bokyung Park,
  Nakjun Sung, Hayeong Yu, Sungjae Kim, SungHyun Park, et~al.
\newblock Surgical scene segmentation using semantic image synthesis with a
  virtual surgery environment.
\newblock In {\em Medical Image Computing and Computer Assisted
  Intervention--MICCAI 2022: 25th International Conference, Singapore,
  September 18--22, 2022, Proceedings, Part VII}, pages 551--561. Springer,
  2022.

\bibitem{yu2019ea}
Biting Yu, Luping Zhou, Lei Wang, Yinghuan Shi, Jurgen Fripp, and Pierrick
  Bourgeat.
\newblock Ea-gans: edge-aware generative adversarial networks for
  cross-modality mr image synthesis.
\newblock {\em IEEE transactions on medical imaging}, 38(7):1750--1762, 2019.

\bibitem{yu2020sample}
Biting Yu, Luping Zhou, Lei Wang, Yinghuan Shi, Jurgen Fripp, and Pierrick
  Bourgeat.
\newblock Sample-adaptive gans: linking global and local mappings for
  cross-modality mr image synthesis.
\newblock {\em IEEE transactions on medical imaging}, 39(7):2339--2350, 2020.

\bibitem{zhang2017stackgan}
Han Zhang, Tao Xu, Hongsheng Li, Shaoting Zhang, Xiaogang Wang, Xiaolei Huang,
  and Dimitris~N Metaxas.
\newblock Stackgan: Text to photo-realistic image synthesis with stacked
  generative adversarial networks.
\newblock In {\em Proceedings of the IEEE international conference on computer
  vision}, pages 5907--5915, 2017.

\bibitem{zhang2019harmonic}
Rui Zhang, Tomas Pfister, and Jia Li.
\newblock Harmonic unpaired image-to-image translation.
\newblock {\em arXiv preprint arXiv:1902.09727}, 2019.

\bibitem{zheng2021spatially}
Chuanxia Zheng, Tat-Jen Cham, and Jianfei Cai.
\newblock The spatially-correlative loss for various image translation tasks.
\newblock In {\em Proceedings of the IEEE/CVF conference on computer vision and
  pattern recognition}, pages 16407--16417, 2021.

\bibitem{zhu2017unpaired}
Jun-Yan Zhu, Taesung Park, Phillip Isola, and Alexei~A Efros.
\newblock Unpaired image-to-image translation using cycle-consistent
  adversarial networks.
\newblock In {\em Proceedings of the IEEE international conference on computer
  vision}, pages 2223--2232, 2017.

\bibitem{zhu2017toward}
Jun-Yan Zhu, Richard Zhang, Deepak Pathak, Trevor Darrell, Alexei~A Efros,
  Oliver Wang, and Eli Shechtman.
\newblock Toward multimodal image-to-image translation.
\newblock {\em Advances in neural information processing systems}, 30, 2017.

\end{thebibliography}
}

\begin{center}
    \textbf{\large Supplementary material: Exploring Semantic Consistency in Unpaired Image Translation to Generate Data for Surgical Applications}
\end{center}
\setcounter{section}{0}
\renewcommand{\thesection}{\Alph{section}}

\section{Dataset}
\begin{figure}
\begin{center}
\includegraphics[width=0.45\textwidth, height=5cm]{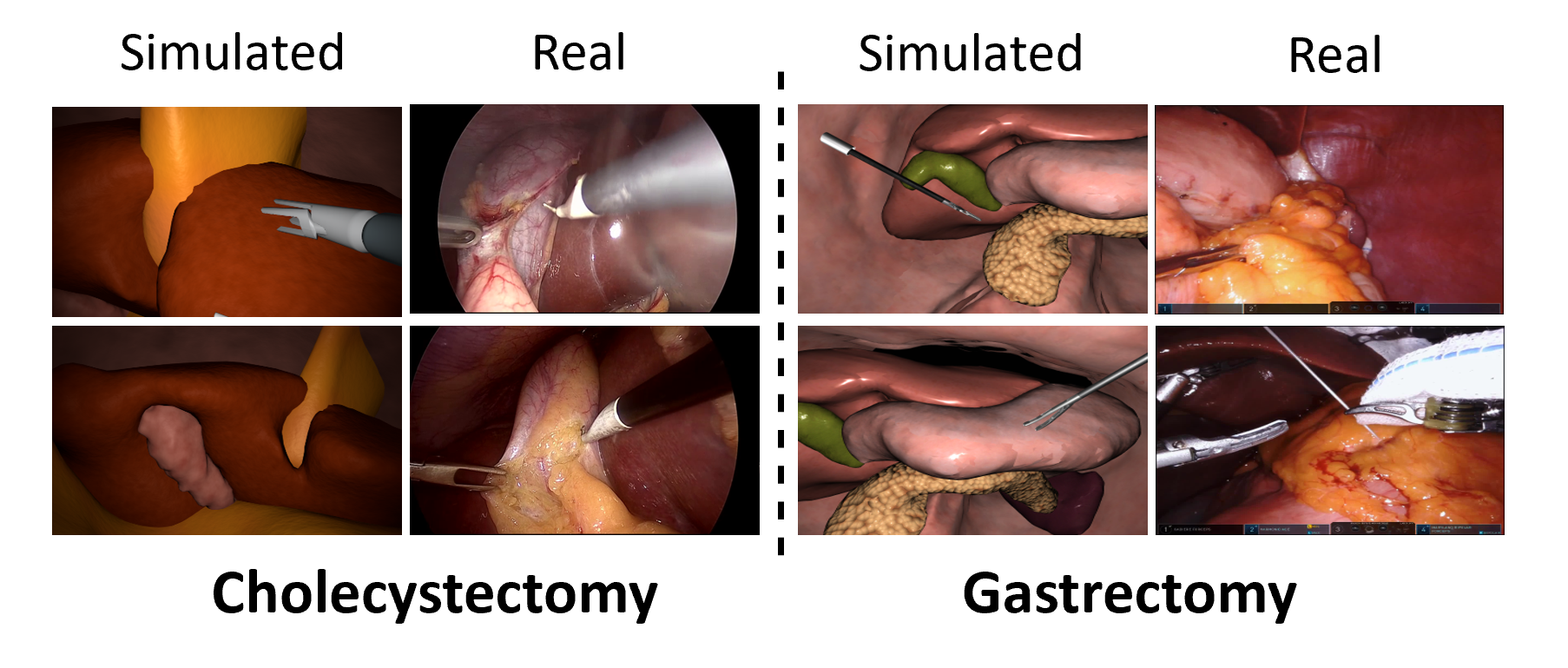}
\end{center}
   \caption{Examples from the surgical datasets.}
\label{fig:supp_dataset}
\end{figure}

For the cholecystectomy dataset, the liver meshes were taken from a public dataset (3D-IRCADb 01 data set, IRCAD, France), while all other structures were designed manually. The camera was moved around along with the light source of the laparoscope, and the synthetic images were rendered. For the real dataset, the images were extracted at a frame rate of five frames per second. A total of $75$ videos were chosen and the images were then curated to remove scenes with only anatomical structures or tools, and finally, a dataset of $26,000$ images was composed. The remaining $5$ videos were chosen fro the \emph{downstream} evaluation. The synthetic dataset has been downloaded from~\url{http://opencas.dkfz.de/image2image/}. Similarly, for the gastrectomy dataset, the entire dataset along with labels has been downloaded from~\url{https://www.kaggle.com/datasets/yjh4374/sisvse-dataset}. Figure~\ref{fig:supp_dataset} shows some examples of the dataset.

\section{Training details}

The architectures and hyperparameters were intentionally matched to have a reasonable performance comparison to the CUT~\cite{park2020contrastive} and CycleGAN~\cite{zhu2017unpaired} models and their variants. Following~\cite{pfeiffer2019generating}, the semantic loss was operated on the images' brightness (average over the channels) as this retains the brightness variations while avoiding the penalization of style-related changes in hue. 

For the ConStructS model, we used a discriminator similar to CUT~\cite{park2020contrastive} but replaced the normalization layers with the spectral norm~\cite{miyato2018spectral} for stabilized training. The Adam optimizer~\cite{kingma2014adam} was utilized with a learning rate of $2e{-4}$ with a linear decay in learning rate. The generator also serves the purpose of the feature encoder to compute the contrastive loss. Correspondingly, the features were encoded from the $1,4,8,12$ and $16^{th}$ layer of the generator. This was constantly maintained for both datasets. The encoded features are passed through a two-layer MLP with $256$ neurons each to extract the feature vectors and are normalized with the $L2$ norm. These feature vectors were utilized for computing the \emph{PatchNCE} loss at $256$ different locations. A batch size of $1$ was employed throughout. For the cholecystectomy dataset, the model was trained for approximately $600$K iterations, whereas for the gastrectomy dataset, the training was carried out until $500$K iterations. The models were trained on single NVIDIA RTX A5000 GPUs with 24GB memory.

All the baseline models were trained based on author's code as mentioned below,
\begin{itemize}
    \item CycleGAN~\cite{zhu2017unpaired}: \url{https://github.com/junyanz/pytorch-CycleGAN-and-pix2pix}
    \item GcGAN~\cite{fu2019geometry}: \url{https://github.com/hufu6371/GcGAN/tree/master}
    \item DistanceGAN~\cite{tran2018dist}: \url{https://github.com/sagiebenaim/DistanceGAN/tree/master}
    \item DRIT++~\cite{lee2018diverse_new}: \url{https://github.com/HsinYingLee/DRIT/}
    \item LapMUNIT~\cite{pfeiffer2019generating}: \url{https://gitlab.com/nct_tso_public/laparoscopic-image-2-image-translation}
    \item UGAT-IT~\cite{kim2019u}: \url{https://github.com/znxlwm/UGATIT-pytorch}
    \item NEGCUT~\cite{wang2021instance}: \url{https://github.com/WeilunWang/NEGCUT}
    \item F/LeSim~\cite{zheng2021spatially}: \url{https://github.com/lyndonzheng/F-LSeSim}
    \item SCC~\cite{guo2022alleviating}: \url{https://github.com/CR-Gjx/SCC}
    \item SRC~\cite{jung2022exploring}: \url{https://github.com/jcy132/Hneg_SRC}
    \item SRUNIT~\cite{jia2021semantically}: \url{https://github.com/SeanJia/SRUNIT}
    \item Perceptual loss~\cite{johnson2016perceptual}: \url{https://github.com/dxyang/StyleTransfer}
    \item CUT~\cite{park2020contrastive}: \url{https://github.com/taesungp/contrastive-unpaired-translation}
\end{itemize}
The same image size was maintained throughout for each dataset during training of all the baseline models.  

\section{Evaluation details}
For the cholecystectomy dataset, the CholecSeg$8$K dataset~\cite{hong2020cholecseg8k}, which is an annotated subset of real images from the Cholec $80$ dataset, was used for training. The training dataset consists of $6000$ images with a test set of $2020$ images. The DeepLabV3+~\cite{chen2018encoder} model was chosen as the segmentation network. We curated the dataset and defined six classes, namely, the liver, abdominal wall, fat, ligament, gallbladder, surgical tools, and background. The evaluation was posed as a multiclass segmentation problem. The different partitions of the tools were fused together into a single tool class. Similarly, for the gastrectomy dataset, six classes were defined: surgical tools, liver, stomach, spleen, pancreas, and gallbladder. Following~\cite{yoon2022surgical}, the models were trained on three different folds of train and test datasets. 

For the \emph{downstream} eval. method, five separate videos were chosen from the Cholec80 dataset. These videos were chosen such that they were not present in the CholecSeg$8$K dataset. The translation models are not exposed these images during training. Since annotating all the tissues and tools would require the guidance of a medical professional and, to simplify the process, only the liver tissue was annotated. The labelme~\cite{labelme} package was used to manually annotate the liver tissue in $196$ images from five different patients. The regions of the liver with minimal lighting were under-segmented in case of doubt to ease the annotation process. A similar DeepLabV3+ model was employed to classify the liver organ. The models were trained and evaluated in leave-one-patient-out method i.e., the model is trained on four patients and evaluated on one patient and this procedure is followed five times. Finally, the mean dice scores is reported. The Adam optimizer was used along with a learning rate of $1e{-5}$. The Onecycle~\cite{smith2019super} scheduler was used to modulate the learning rate during training. 
\section{Additional results}
\begin{figure*}
\begin{center}
\includegraphics[width=0.95\textwidth]{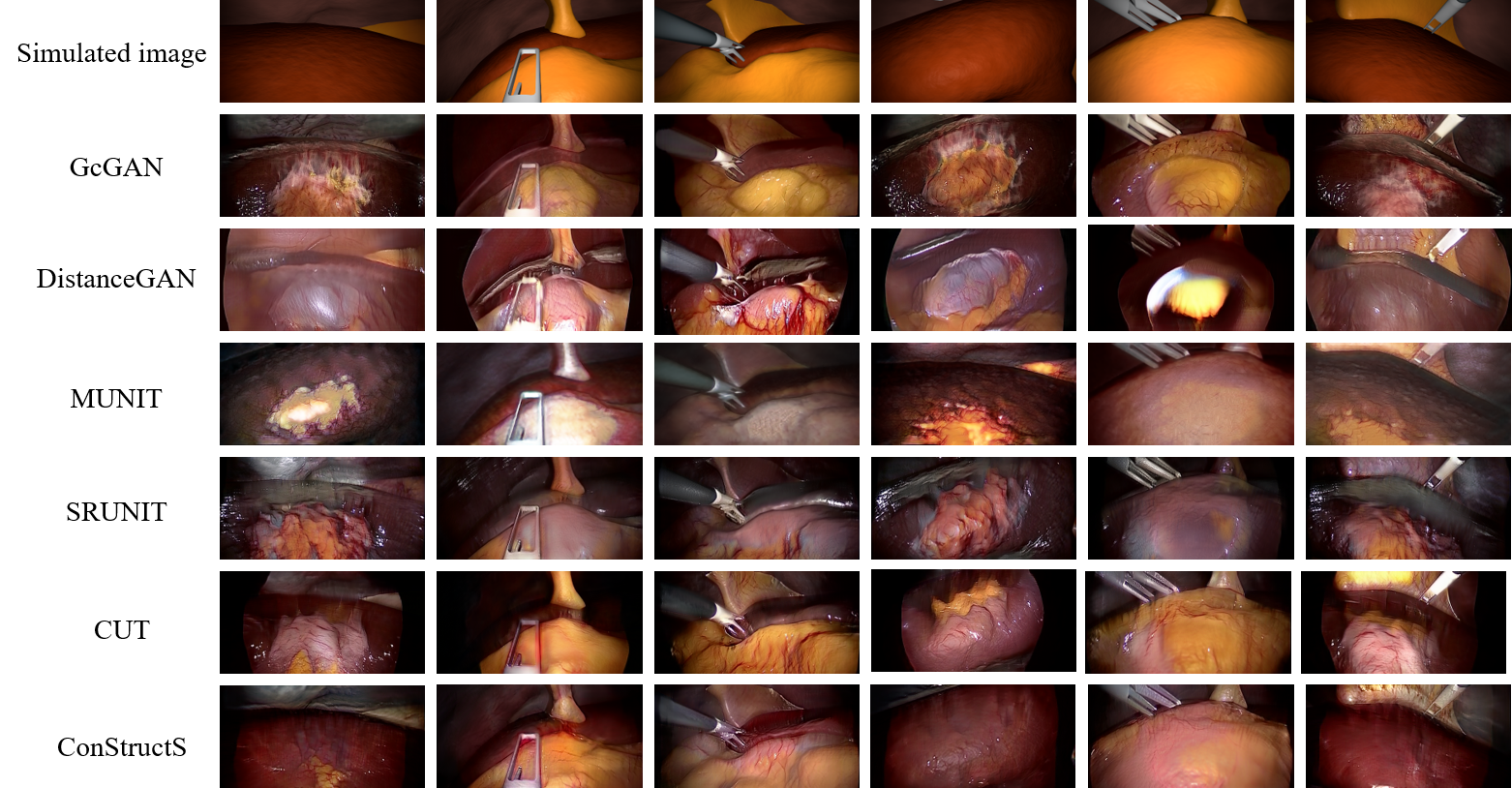}
\end{center}
   \caption{Additional results from the cholecystectomy dataset.}
\label{fig:supp_cholec}
\end{figure*}
\subsection{Cholecystectomy} 
\begin{figure*}
\begin{center}
\includegraphics[width=0.95\textwidth]{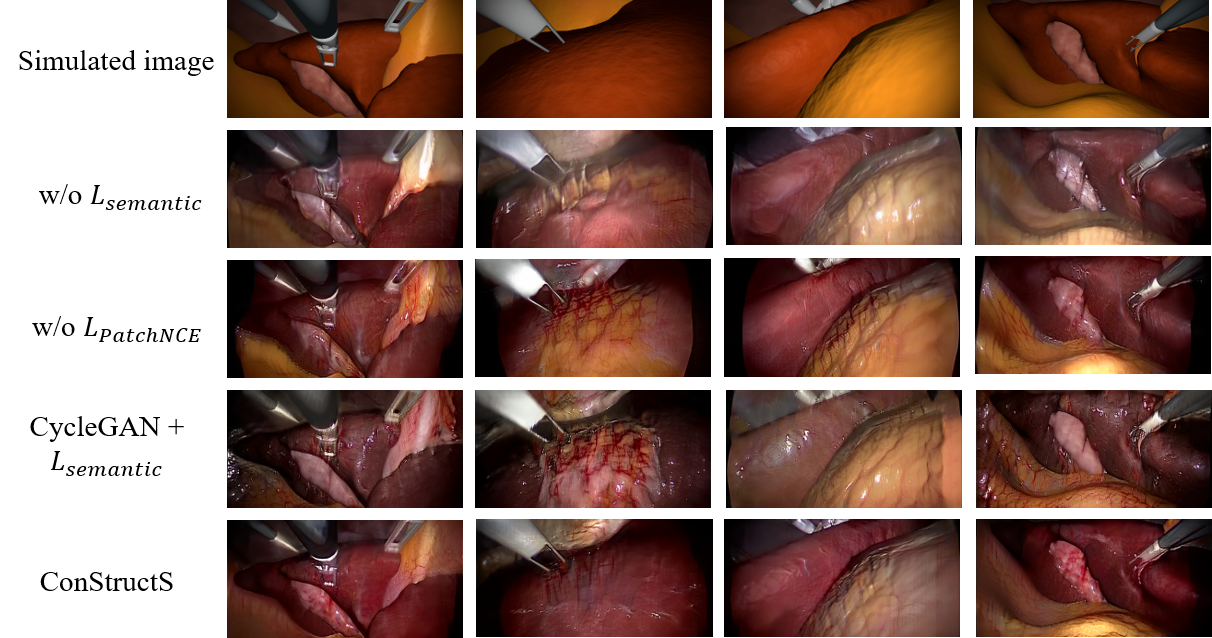}
\end{center}
   \caption{Additional results from the ablation study on the cholecystectomy dataset.}
\label{fig:supp_ablation}
\end{figure*}

We provide additional qualitative results for both datasets. The Figure~\ref{fig:supp_cholec} further indicates that the ConStructS model is able to maintain both structure and semantic consistency when compared to many of the other models. Furthermore, the additional results of the ablation study from Figure~\ref{fig:supp_ablation} show the importance of combining the \emph{PatchNCE} loss with the \emph{semantic} loss to reduce semantic distortion.  Table~\ref{tab:cont} indicates the results for an ablation study where the ConStructS method is trained with and without the $\mathcal{L}_{Patch}(Y)$. The model performance deteriorates in the absence of $\mathcal{L}_{Patch}(Y)$.

\begin{table}
  \begin{center}
    {\small{
    \resizebox{\linewidth}{!}{
\begin{tabular}{lccc}
\toprule
Models & pxAcc & clsAcc & mIOU \\
\midrule
ConStructS w/o $\mathcal{L}_{Patch}(Y)$& $0.50 \pm 0.06$ & $0.41 \pm 0.13$ & $0.25 \pm 0.08$ \\
ConStructS & $\mathbf{0.59 \pm 0.07}$ & $\mathbf{0.44 \pm 0.12}$ & $\mathbf{0.29 \pm 0.09}$ \\
\bottomrule
\end{tabular}
}
}}
\end{center}
\caption{The \emph{consistency} evaluation results with and without the $\mathcal{L}_{Patch}(Y)$ loss.}
\label{tab:cont}
\end{table}

\subsection{Gastrectomy} 
\begin{figure*}
\begin{center}
\includegraphics[width=0.95\textwidth]{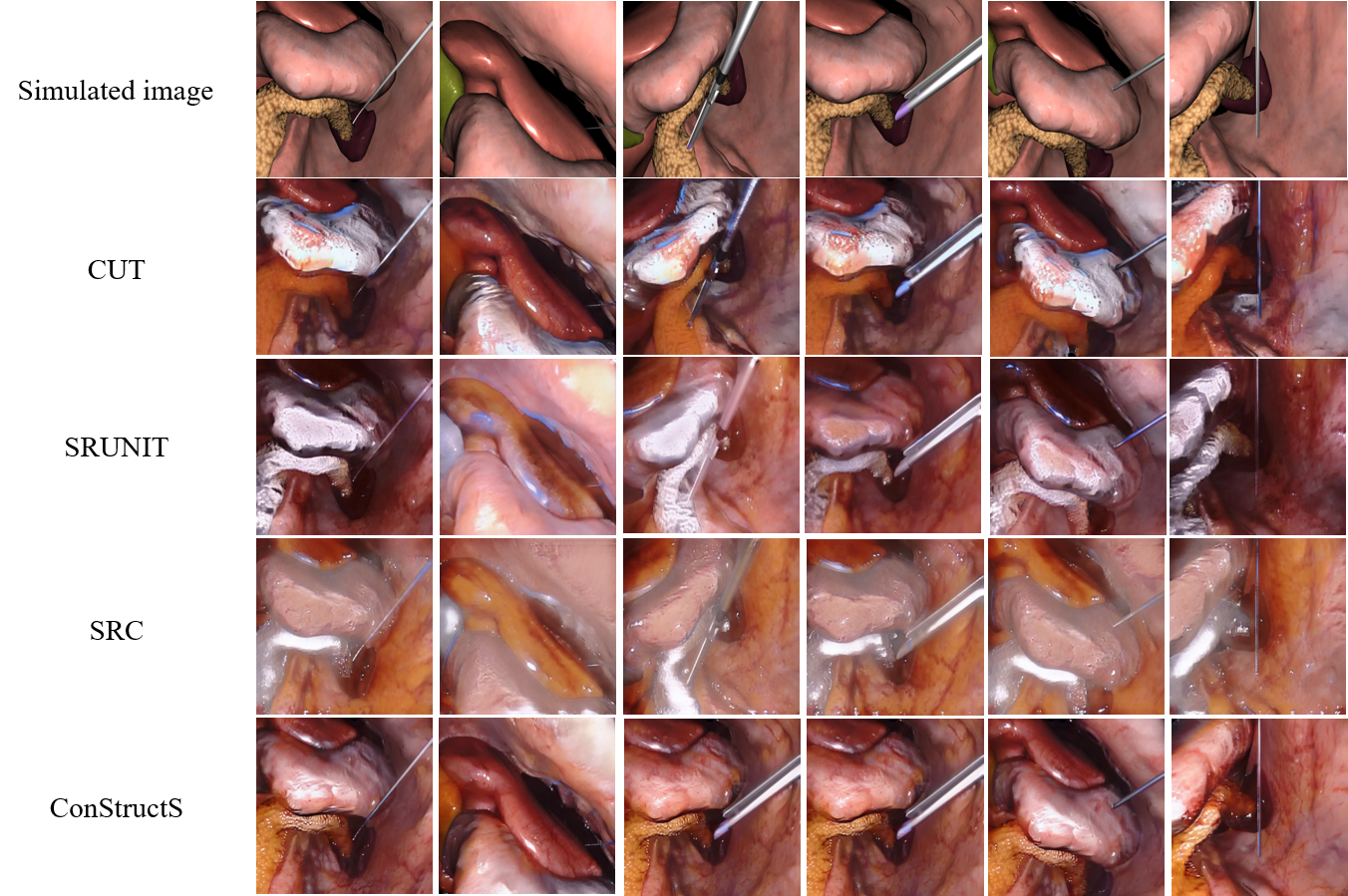}
\end{center}
   \caption{Additional results from the the gastrectomy dataset.}
\label{fig:supp_gast}
\end{figure*}

The visual results depicted in Figure~\ref{fig:supp_gast} demonstrate that ConStructS significantly mitigates semantic mismatches, particularly in regions characterized by differing specularity compared to other models. For the gastrectomy dataset, we noticed that the real images contain three extra classes compared to the synthetic domain. Mainly, there existed a class/element gauze(white) that exceeded in proportion to the other classes. This semantic imbalance was reflected in the translation performance, where many models mapped this texture, especially to regions with higher specularity.

\subsection{Downstream evaluation}
In Table~\ref{tab:multi_acc} the segmentation scores of various models are shown for multi-class downstream evaluation. The translated images from the LapMUNIT~\cite{pfeiffer2019generating} outperforms most the SOTA models. However, the images from the ConStructS model proves useful in improving the scores upto $6\%$. In Table~\ref{tab:supp_eval2}, the results on \emph{eval-$2$} method is reported. When compared to the baseline models, using only the translated images as training data from ConStructS leads to an overall improvement of $9\%$ in dice scores

\subsection{Metrics} We report FID score for the different models on the cholecystectomy dataset. The FID values for both the datasets on various layers are shown in Table~\ref{tab:supp_fid} and Table~\ref{tab:supp_fid_gast} respectively. The general practice is to indicate the values on the layer with $2048$ features. However, here we find an effect that different layers show different values. This is one of the reasons for the adoption of different evaluation schemes in this study.

\begin{table}
  \begin{center}
    {\small{
    \resizebox{\linewidth}{!}{
\begin{tabular}{lccccc}
\toprule
Training data & \multicolumn{2}{c}{mean dice}\\

& Pre-train & Fine-tune on real \\
\midrule
Baseline & - & $0.62\pm0.11$\\ \hdashline
CycleGAN$+$VGG~\cite{zhu2017unpaired}& $0.65\pm 0.85$ & $0.73\pm0.08$\\
DRIT$++$~\cite{lee2018diverse_new} & $0.38\pm 0.06$ & $0.59\pm0.10$\\
UGAT-IT~\cite{kim2019u} & $0.36\pm 0.07$ & $0.57\pm0.09$\\ \hdashline
DistGAN~\cite{tran2018dist} & $0.36\pm 0.01$ & $0.58\pm0.06$\\ \hdashline
NEGCUT~\cite{wang2021instance}& $0.65\pm 0.10$ & $0.72\pm0.09$\\
FeSim~\cite{zheng2021spatially}& $0.19\pm 0.90$ & $0.27\pm0.10$\\
LeSim~\cite{zheng2021spatially}& $0.45\pm 0.17$ & $0.72\pm0.10$\\ \hdashline
CycleGAN$+$SCC~\cite{guo2022alleviating} & $0.56\pm 0.20$ & $0.73\pm0.10$\\
CUT$+$SCC~\cite{guo2022alleviating} & $0.38\pm 0.15$ & $0.63\pm0.08$\\ \hdashline
CUT~\cite{park2020contrastive} & $0.60\pm 0.12$ & $0.71\pm0.02$\\
ConStructS & $0.65\pm0.21$ & $0.84\pm0.05$\\
\bottomrule
\end{tabular}
}
}}
\end{center}
\caption{The quantitative results for downstream eval. The mean dice scores are reported.}
\label{tab:supp_eval2}
\end{table}

\begin{figure*}
\begin{center}
\includegraphics[width=0.95\textwidth]{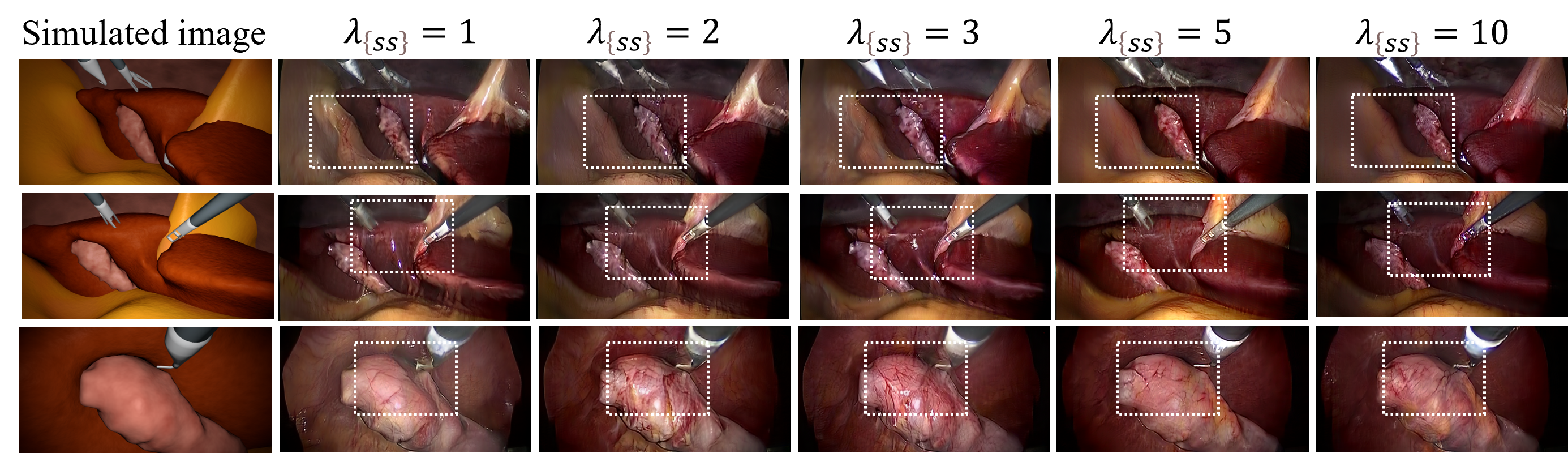}
\end{center}
   \caption{Additional results from the sensitivity analysis on the cholecystectomy dataset. The structure of the surgicall tool is the best maintained with $\lambda_{SS}=5$. Similarly, for the same value we find that blood texture is not mixed with either the liver organ or the junction between the ligament and liver.}
\label{fig:supp_sens}
\end{figure*}

\begin{table}
  \begin{center}
    {\small{
    \resizebox{\linewidth}{!}{
\begin{tabular}{lcccc}
\toprule
Method & L$64$ & L$192$ & L$768$ & L$2048$ \\
\midrule
CycleGAN~\cite{zhu2017unpaired}& $2.21$ & $8.17$ & $0.78$ & $136.82$ \\
GcGAN~\cite{lee2018diverse_new} & $2.04$ & $9.23$ & $0.62$ & $127.06$ \\
LapMUNIT~\cite{kim2019u} & $0.73$ & $3.92$ & $0.82$ & $167.48$ \\ 
CUT~\cite{tran2018dist} & $1.88$ & $9.05$ & $0.65$ & $141.31$ \\
CUT$+$SCC~\cite{guo2022alleviating} & $2.89$ & $10.02$ & $0.68$ & $126.25$ \\
SRUNIT~\cite{wang2021instance}& $3.41$ & $10.36$ & $0.79$ & $138.87$ \\
SRC~\cite{zheng2021spatially}& $2.02$ & $7.79$ & $0.71$ & $134.82$\\ \hdashline
ConStructS & $1.29$ & $6.70$ & $0.69$ & $121.03$\\
\bottomrule
\end{tabular}
}
}}
\end{center}
\caption{The FID scores for the translation models on cholecystectomy dataset.}
\label{tab:supp_fid}
\end{table}

\begin{table}
  \begin{center}
    {\small{
    \resizebox{\linewidth}{!}{
\begin{tabular}{lcccc}
\toprule
Method & L$64$ & L$192$ & L$768$ & L$2048$ \\
\midrule
CycleGAN~\cite{zhu2017unpaired}& $3.10$ & $12.76$ & $0.82$ & $200.64$ \\
GcGAN~\cite{lee2018diverse_new} & $1.42$ & $5.86$ & $0.84$ & $193.52$ \\
LapMUNIT~\cite{kim2019u} & $1.26$ & $5.55$ & $0.82$ & $183.27$ \\ 
CUT~\cite{tran2018dist} & $1.75$ & $7.40$ & $0.89$ & $192.11$ \\
SRUNIT~\cite{wang2021instance}& $3.22$ & $11.63$ & $0.75$ & $207.77$ \\
SRC~\cite{zheng2021spatially}& $6.20$ & $24.29$ & $0.71$ & $196.91$\\ \hdashline
ConStructS & $2.08$ & $9.95$ & $0.77$ & $171.60$\\
\bottomrule
\end{tabular}
}
}}
\end{center}
\caption{The FID scores for the translation models on gastrectomy dataset.}
\label{tab:supp_fid_gast}
\end{table}

\begin{table}
  \begin{center}
    {\small{
    \resizebox{\linewidth}{!}{
\begin{tabular}{lcccccc}
\toprule
Models & pxAcc & clsAcc & mIOU \\
\midrule
Baseline & $0.82$ & $0.75$ & $0.74$\\
CycleGAN~\cite{zhu2017unpaired}& $0.84$ & $0.76$ & $0.76$\\
GcGAN~\cite{lee2018diverse_new} & $0.83$ & $0.74$ & $0.74$\\
LapMUNIT~\cite{kim2019u} & $0.85$ & $0.76$ & $0.74$\\ 
CUT~\cite{tran2018dist} & $0.83$ & $0.75$ & $0.75$\\
SRUNIT~\cite{wang2021instance}& $0.81$ & $0.73$ & $0.75$\\
ConStructS & $0.88$ & $0.76$ & $0.78$ \\
\bottomrule
\end{tabular}
}
}}
\end{center}
\caption{The multi-class segmentation scores for the \emph{downstream} evaluation.}
\label{tab:multi_acc}
\end{table}

\end{document}